\documentclass[runningheads]{llncs}

\usepackage{eccv}

\usepackage{authblk}
\usepackage{marvosym}

\definecolor{myblue}{rgb}{0.21,0.49,0.74}
\usepackage[pagebackref,breaklinks,colorlinks,allcolors=myblue]{hyperref}

\usepackage{eccvabbrv}

\usepackage{graphicx}
\usepackage{booktabs}

\usepackage[accsupp]{axessibility}  

\usepackage{hyperref}

\usepackage{orcidlink}
\newcommand{\ourbench}{\textsc{BizGenEval}\xspace}

\usepackage{multirow}
\usepackage[table,xcdraw]{xcolor}
\usepackage[normalem]{ulem}
\useunder{\uline}{\ul}{}

\usepackage{graphicx}
\usepackage{subcaption}

\usepackage{xcolor}
\usepackage{wrapfig}
\usepackage{tabularx}

\usepackage{tcolorbox} 
\usepackage{caption} 

\newtcolorbox{promptbox}[1][]{
  colback=gray!10!white, 
  colframe=gray!50!black, 
  boxrule=0.5pt, 
  width=\linewidth, 
  sharp corners, 
  before=\vspace{1em}, 
  after=\vspace{0.5em}, 
  left=1mm, right=1mm, top=1mm, bottom=1mm, 
  #1
}

\definecolor{hard_mode}{RGB}{0, 123, 167}
\definecolor{easy_mode}{RGB}{147, 143, 139}
\newcommand{\score}[2]{#2/\textcolor{easy_mode}{#1}}

\newcommand{\scor}[2]{\makebox[2.5em][r]{#2}\,/\,\textcolor{easy_mode!60}{#1}}

\definecolor{openmodel}{RGB}{235,245,255}   
\definecolor{closedmodel}{RGB}{204, 204, 255} 
\newcommand{\openmodel}[1]{\cellcolor{openmodel}#1}
\newcommand{\closedmodel}[1]{\cellcolor{closedmodel!40}#1}

\definecolor{darkblue}{RGB}{61,89,171}
\definecolor{darkred}{RGB}{178,34,34}

\begin{document}

\title{\ourbench: A Systematic Benchmark for Commercial Visual Content Generation} 

\titlerunning{\ourbench{}}

\author{
Yan Li\inst{2}$^{*}$ \and
Zezi Zeng\inst{3}$^{*}$ \and
Ziwei Zhou\inst{4}$^{*}$ \and
Xin Gao\inst{4}$^{*}$ \and
Muzhao Tian\inst{4}$^{*}$ \and
Yifan Yang\inst{1}$^{\dagger}$ \and
Mingxi Cheng\inst{1} \and
Qi Dai\inst{1} \and
Yuqing Yang\inst{1} \and
Lili Qiu\inst{1} \and
Zhendong Wang\inst{1} \and
Zhengyuan Yang\inst{1} \and
Xue Yang\inst{2}$^{\dagger}$ \and
Lijuan Wang\inst{1} \and
Ji Li\inst{1} \and
Chong Luo\inst{1}
}

\authorrunning{Y. Li et al.}

\institute{
Microsoft Corporation
\and
Shanghai Jiao Tong University
\and
Xi'an Jiaotong University
\and
Fudan University
}

\maketitle

\begin{center}
\small \url{https://aka.ms/BizGenEval}
\end{center}

\begingroup
\renewcommand\thefootnote{}
\footnotetext{$^{*}$ Equal contribution. This work was done during their internship at Microsoft.}
\footnotetext{$^{\dagger}$ Corresponding to: Yifan Yang <yifanyang@microsoft.com>, Xue Yang <yangxue-2019-sjtu@sjtu.edu.cn>.}
\endgroup

\begin{abstract}
Recent advances in image generation models have expanded their applications beyond aesthetic imagery toward practical visual content creation. However, existing benchmarks mainly focus on natural-image synthesis and fail to systematically evaluate models under the structured and multi-constraint requirements of real-world commercial design tasks. In this work, we introduce \ourbench{}, a systematic benchmark for commercial visual content generation. The benchmark spans five representative document types—slides, charts, webpages, posters, and scientific figures—and evaluates four key capability dimensions: text rendering, layout control, attribute binding, and knowledge-based reasoning, forming 20 diverse evaluation tasks. \ourbench{} contains 400 carefully curated prompts and 8,000 human-verified checklist questions to rigorously assess whether generated images satisfy complex visual and semantic constraints. We conduct large-scale benchmarking on 26 popular image generation systems, including state-of-the-art commercial APIs such as Nano Banana 2 and GPT-Image-1.5 as well as leading open-source models. The results reveal substantial capability gaps between current generative models and the requirements of professional visual content creation. We hope \ourbench{} serves as a standardized benchmark for real-world commercial visual content generation.


\end{abstract}

\section{Introduction}
\label{sec:intro}

Recent advances in large-scale image generation models, such as Nano Banana Pro~\cite{nanobananapro} and GPT-Image-1.5~\cite{openai2025gptimage1.5}, have significantly expanded the capabilities of visual generation systems. 
Beyond producing aesthetically pleasing images, these models are increasingly being used as practical interfaces for complex visual content creation tasks. 
In many real-world scenarios, image generation systems can already produce professional materials such as presentation slides, web page layouts, scientific figures, posters, and data charts with minimal human intervention. 
Recent industry reports, including GPT-Image-1.5 and Qwen-Image 2, increasingly highlight commercial design scenarios, reflecting the growing practical and economic importance of such capabilities.

However, systematic evaluation of commercial visual generation remains largely under-explored. 
Despite the growing practical importance of these tasks, there is currently no comprehensive benchmark that systematically evaluates how well image generation models perform on commercial visual content creation, and model capabilities are often demonstrated through selected examples as in~\cite{openai2025gptimage1.5,Seedream5,qwenimage2,glm_image,LongCat-Image} rather than standardized evaluation.

\begin{figure}[!htbp]
    \centering
    \includegraphics[width=\linewidth]{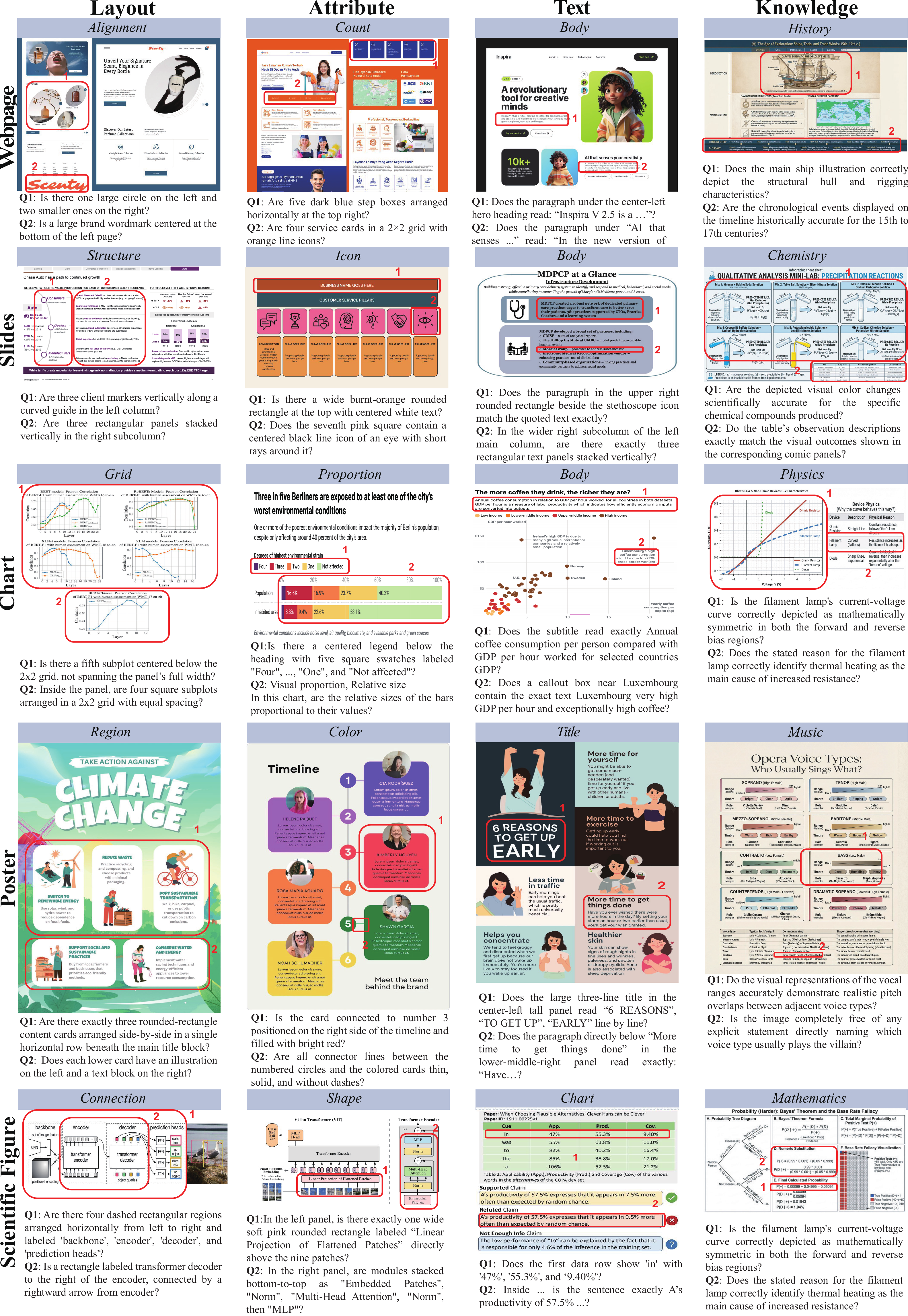}
    \caption{\textbf{Real-world samples of \ourbench from 5 content domains and 4 capability dimensions.} 
    For each capability dimension, different subtasks are highlighted with blue boxes. Each example covers distinct aspects, with the relevant regions highlighted in red and the corresponding simplified questions listed below.}
    \label{fig:data_sample_overview}
\end{figure}

Evaluating such tasks is inherently challenging, as commercial visual documents typically contain dense textual content, complex layouts, numerous visual elements, and strict semantic correctness requirements. 
Existing image generation benchmarks~\cite{ghosh2023geneval,huang2023t2i,DPG-Bench} mainly focus on natural images, object compositionality, or aesthetic quality. OneIG-Bench~\cite{chang2025oneig} includes themes such as slides or poster generation, but mainly uses them as prompt sources for text-rendering evaluation with simple prompts and similarity-based metrics, rather than assessing holistic commercial visual generation capability.
Although several recent works explore design-oriented tasks such as infographic generation~\cite{tang2026igenbench} or poster rendering~\cite{peng2025bizgen}, these evaluations remain fragmented and domain-specific.

To address this gap, we introduce \ourbench{}, the first comprehensive benchmark for evaluating image generation models on commercial visual content creation. 

The benchmark covers \textbf{5} representative domains—\textit{Slides}, \textit{Scientific Figures}, \textit{Charts}, \textit{Webpages}, and \textit{Posters}—which correspond to common visual artifacts in professional workflows. For each domain, we evaluate \textbf{4} key capabilities: \textit{Text Rendering}, \textit{Layout Control}, \textit{Attribute Binding}, and \textit{Knowledge-based Reasoning}. Together, the five domains and four capability dimensions form \textbf{20 evaluation tasks}. Representative examples are shown in Fig.~\ref{fig:data_sample_overview}, illustrating realistic multi-constraint generation scenarios in professional design workflows.

For the \textbf{content-based dimensions} (\textit{Text Rendering}, \textit{Layout Control}, and \textit{Attribute Binding}), prompts are constructed from real-world commercial materials. We manually collect authentic design examples from professional online sources and curate representative samples for each domain. Based on these references, task-specific prompts are created to simulate realistic requests that a human designer might issue when using an image generation system. These prompts test whether models can accurately control low-level design elements such as typography, arrow directions, text box placement, visual counts, and spatial arrangements.

Beyond element-level rendering, commercial visual generation often requires incorporating external domain knowledge. 
To evaluate this aspect, we introduce the \textit{Knowledge-based Reasoning} dimension covering five themes—\textit{physics}, \textit{chemistry}, \textit{mathematics}, \textit{history}, and \textit{arts}. 
These tasks require models to generate coherent images while reflecting underlying factual or conceptual knowledge.

All prompts and evaluation rationales are carefully curated and rewritten by human experts to ensure task validity and difficulty. In total, the benchmark contains \textbf{300 content-based prompts} and \textbf{100 knowledge-based prompts}, enabling comprehensive evaluation of both visual design control and knowledge-grounded generation in realistic commercial content creation scenarios.

Evaluating commercial visual documents is challenging due to their dense textual content, complex layouts, and tightly coupled visual constraints. To address this, we design a structured checklist-based evaluation protocol. For each prompt, human experts construct 20 task-specific verification questions that directly correspond to the evaluated capability.

We further introduce an automated evaluation pipeline that employs state-of-the-art multimodal large language models (MLLMs) as judges to answer these verification questions. The final score is computed based on the accuracy of the responses, enabling scalable and consistent evaluation. Human evaluation in Sec.~\ref{sec:main_results} shows a strong agreement between MLLM evaluators and human judgments, demonstrating the reliability of the proposed protocol.

Using \ourbench{}, we conduct a large-scale evaluation of both closed-source and open-source image generation models. The results reveal substantial performance differences across domains and capability dimensions, providing new insights into the strengths and limitations of current image generation systems for commercial visual content creation.

\textbf{Contributions.}
1) \ourbench{}, the first comprehensive benchmark for commercial visual content generation, covering five representative domains and four capability dimensions, resulting in 20 evaluation tasks and 400 curated prompts, together with 8,000 human-verified checklist questions for rigorous evaluation. 
2) We conduct a large-scale evaluation of state-of-the-art image generation systems, spanning both commercial APIs and open-source models, establishing strong baselines for commercial visual generation. 
3) We provide extensive empirical analysis of current model capabilities and limitations, offering insights for future image generation systems in practical commercial applications.

\section{Related Work}
\label{sec:related_work}
\noindent\textbf{Benchmarks for Commercial Visual Documents.}
Recent advances in large-scale image generation models have increasingly sparked interest in their applications to commercial visual generation, motivating several works to benchmark specific types of commercial visual documents. For example, SlidesGen-Bench~\cite{yang2026slidesgenbench} is restricted to language-driven slide creation, while IGenBench~\cite{tang2026igenbench} and BizGen~\cite{peng2025bizgen} focus on infographics with atomic verification and long-context layout constraints.
For web interfaces, Design2Code~\cite{si2024design2code} and WebSight~\cite{laurencon2024websight} are designed specifically for benchmarking screenshot-to-code systems.
In scientific figures, FigureBench~\cite{zhu2026autofigure} is tailored to scientific figure generation from long-form technical descriptions. Unlike these domain-specific efforts, our \ourbench provides a holistic perspective across five representative commercial domains, covering slides, charts, webpages, posters, and scientific figures.

\noindent\textbf{Benchmarks for Core Image-Generation Capabilities.}
Beyond domain-specific applications, a parallel line of research focuses on evaluating the fundamental generative capabilities of text-to-image models.
For text rendering, LongText-Bench~\cite{geng2025xomni} and TextCrafter with CVTG-2K~\cite{du2025textcrafter} are restricted to long-form or multi-region text on text-centric images. 
Instruction-following suites such as TIIF-Bench~\cite{wei2025tiif} and OneIG-Bench~\cite{chang2025oneig} include a text-rendering dimension, but evaluate it via automatic global semantic scores on relatively unconstrained scenes, and offer little coverage of dense, layout-aware typography in structured commercial documents.
For spatial layout and attribute control, LayoutBench~\cite{cho2023layoutbench}, 7Bench~\cite{izzo2025sevenbench}, and OverLayBench~\cite{li2025overlaybench} operate on abstract bounding-box layouts or simplified layout-to-image settings, with layout specified by object boxes and category labels. They rarely couple layout with long textual content or complex document semantics. GenEval~\cite{ghosh2023geneval}, VQAScore~\cite{lin2024vqascore}, and GenAI-Bench~\cite{li2024genaibench} further probe object-level attributes, counts, and compositional relations in natural or synthetic scenes, rather than design-centric properties such as color schemes, icon usage, and stylistic consistency inside structured commercial documents.
For knowledge and reasoning, WISE~\cite{niu2025wise}, WorldGenBench~\cite{zhang2025worldgenbench}, R2I-Bench~\cite{chen2025r2ibench}, and MMMG~\cite{luo2025mmmg} evaluate knowledge-grounded image synthesis on generic or educational-style scenes, typically without enforcing realistic document formats or multi-constraint design goals. GenExam~\cite{wang2025genexam} introduces exam-style prompts and checklist scoring, but still operates outside commercial document layouts and does not require simultaneous satisfaction of layout, dense text, attributes, and factual correctness.
\section{\ourbench}

\subsection{Task Description}

\ourbench is designed to evaluate commercial visual generation across diverse real-world content creation scenarios and demanding capabilities. 
Specifically, the benchmark is structured along two orthogonal dimensions: \textbf{5} content domains and \textbf{4} capability dimensions, resulting in a total of \textbf{20} evaluation tasks, as illustrated in Fig.~\ref{fig:data_sample_overview}. 

We first start with five representative types of \textit{\textbf{commercial visual content}} that frequently arise in professional workflows:

\noindent
\textbf{Webpage}: webpage designs that integrate structured layouts, textual content, and functional interface elements such as headers, sections, buttons, etc.

\noindent
\textbf{Slides}:  presentation slides used in reports, lectures, and business scenarios, characterized by hierarchical structure, bullet lists, and aligned visual elements.

\noindent
\textbf{Chart}: data visualization graphics such as  bar charts, line charts, and multi-series plots, involving precise rendering of numeric values, axes,  and legends.

\noindent
\textbf{Poster}: promotional or informational posters combining typography, graphics, and layout design, often emphasizing visual hierarchy and balanced composition.

\noindent
\textbf{Scientific Figure}: figures in academic papers including diagrams, pipelines, and illustrations, where components, arrows, and annotations form clear structure.

For each domain, we further design four \textbf{\textit{capability-oriented evaluation tasks}} that capture key challenges in commercial visual generation:

\noindent
\textbf{Layout}: evaluates spatial and structural organization, including overall layout, complex flows, arrows, blocks, and hierarchical arrangement of elements.

\noindent
\textbf{Attribute}: assesses control over visual attributes such as color, shape, style, icons, and quantity, emphasizing fine-grained control.

\noindent
\textbf{Text}: evaluates textual content rendering, including short titles, long paragraphs, tables, and integration with other components.

\noindent
\textbf{Knowledge}: tests reasoning and domain knowledge, including applying world knowledge across diverse domains such as physics, chemistry, arts, history, etc.

\begin{figure}[!htbp]
    \centering
    \includegraphics[width=\linewidth]{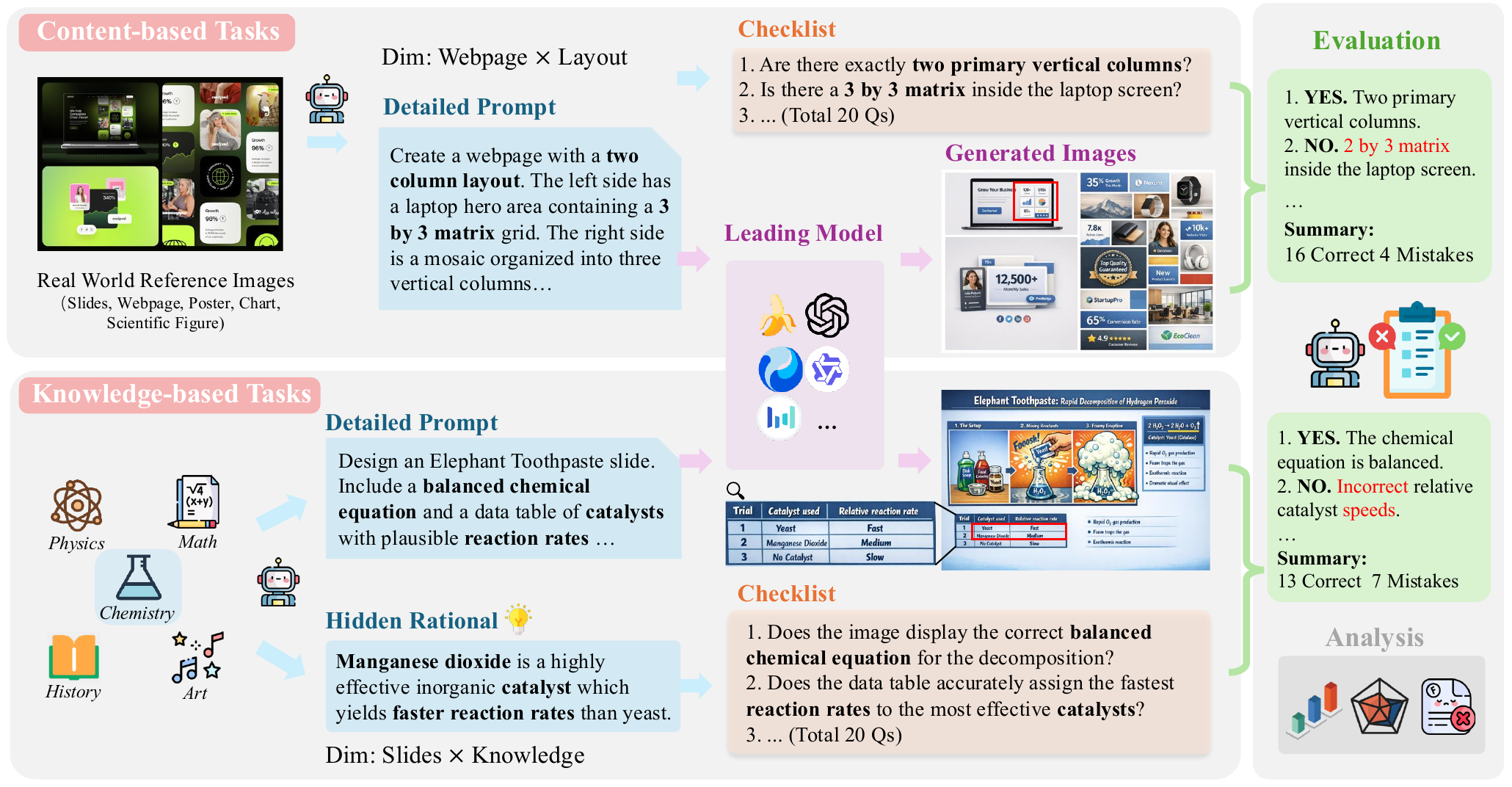}
    \caption{\textbf{Overview of the construction and evaluation pipeline.} The system converts real world references and domain knowledge into structured prompts, and evaluates generated images using rigorous checklists to provide a comprehensive analysis.}
    \label{fig:pipeline}
\end{figure}

\subsection{Benchmark Construction of \ourbench}

\noindent\textbf{Content-based Data Collection.}
To support evaluation of the content-based capabilities (\textit{Text Rendering}, \textit{Layout Control}, and \textit{Attribute Binding}), we first construct a reference pool of real-world commercial visual designs. We manually collect \textbf{1,819} candidate images from diverse professional sources, including UI/UX design repositories, corporate presentation archives, academic databases, and digital marketing portfolios. Unlike aesthetic-oriented datasets, our collection explicitly focuses on visuals created for authentic commercial scenarios.

Based on the predefined \textit{domain–task} taxonomy, we then perform a targeted curation process to identify representative references for each domain–capability combination. For every domain–task pair, we deliberately select images that exhibit clear structural patterns and realistic design constraints. Through multiple rounds of \textit{human-in-the-loop} filtering, low-information or ambiguous samples are excluded while any personal or sensitive information in the original materials is removed to ensure privacy compliance.

Finally, we retain \textbf{20} representative references for each domain–task combination, ensuring balanced coverage across domains and capabilities. These curated references later serve as the foundation for prompt construction. In total, this process produces \textbf{300} content-based reference samples capturing key challenges in commercial visual content generation.

\noindent\textbf{Knowledge-based Data Collection.}
Beyond structural rendering, commercial visual generation often requires models to incorporate domain knowledge and factual reasoning. To evaluate this capability, we construct a knowledge reference pool covering five themes: \textit{physics}, \textit{chemistry}, \textit{mathematics}, \textit{history}, and \textit{arts}.
For each theme, \textbf{20} representative knowledge points are curated from professional databases, educational resources, and academic curricula to ensure that the evaluation targets reflect reliable and expert-level content. For scientific domains such as \textit{physics} and \textit{chemistry}, we prioritize knowledge points involving experimental setups, molecular structures, and fundamental scientific principles that require precise symbolic correctness and conceptual consistency. The \textit{mathematics} category emphasizes reasoning-based visual tasks such as geometric constructions, equations, and diagrammatic representations. For humanities-related themes including \textit{history} and \textit{arts}, the selected knowledge points focus on chronological relationships, cultural symbols, and stylistic characteristics.

Through this process we obtain \textbf{100} curated knowledge references. These knowledge points are later expanded into generation prompts across the benchmark domains to evaluate models' ability to integrate factual knowledge into visual content generation.

\noindent\textbf{Prompt Construction.}
Based on the curated visual references and knowledge points, we construct prompts for both content-based and knowledge-based evaluation tasks. For \textit{\textbf{content-based tasks}}, prompts are generated from reference images using a structured analysis pipeline. A vision–language model performs component-level analysis to identify key structural elements, visual attributes, and textual content, from which detailed generation instructions are produced to specify the required visual elements and their relationships. Different tasks emphasize distinct aspects of generation: \textit{layout control} focuses on spatial organization and structural relationships, \textit{attribute binding} targets visual properties such as color, shape, quantity, and style, and \textit{text rendering} requires exact character-level reproduction of textual content.

For \textit{\textbf{knowledge-based tasks}}, prompts are constructed by expanding curated knowledge references into generation instructions within the five content domains. To prevent answer leakage and encourage genuine reasoning, key knowledge facts are intentionally omitted from the prompts and retained as hidden rationales used only during evaluation. For example, as illustrated in Fig.~\ref{fig:pipeline}, a task combining the Slides domain and Chemistry knowledge asks the model to \textit{``Design an Elephant Toothpaste slide with a balanced chemical equation and a catalyst table with plausible reaction rates.''} The hidden rationale that \textit{``Manganese dioxide is a highly effective inorganic catalyst that yields faster reaction rates than yeast''} is withheld, forcing the model to rely on its internal knowledge to produce factually consistent visual content.

\begin{figure}[!htbp]
    \centering
    \includegraphics[width=\linewidth]{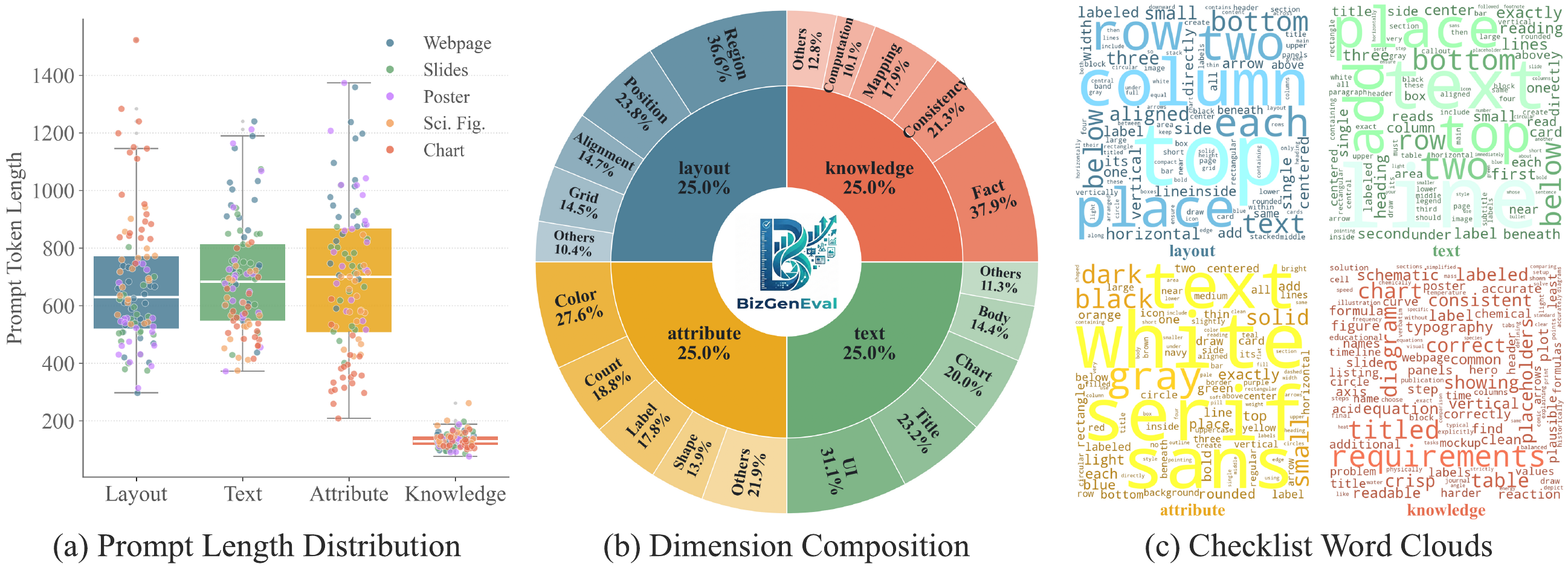}
    \caption{\textbf{Dataset Statistics of \ourbench{}.}
(a) Prompt token length distribution by evaluation dimension, with scatter points color-coded by document type.
(b) Hierarchical subcategory distribution across evaluation dimensions. The inner ring shows question proportions per dimension; the outer ring shows subcategory breakdowns.
(c) Checklist Questions keyword clouds per evaluation dimension.}
    \label{fig:data_statistics}
\end{figure}

\noindent\textbf{Evaluation Checklist Construction.}
Based on the prompt, we further construct \textbf{20} verification questions for each task through a \textit{human-in-the-loop} process, forming a structured checklist for systematic evaluation. Due to the large variability in model performance, we divide the questions into \textbf{10 easy} and \textbf{10 hard} items to enhance discrimination across models of different capability levels. The easy questions mainly target fundamental properties such as basic spatial layout, prominent visual attributes, or high-level text correctness, while the hard questions require fine-grained reasoning and precise control over visual, structural, or textual elements.

\noindent\textbf{Manual Verification.}
To ensure the reliability and quality of the benchmark, we conduct a multi-round manual verification process over the generated prompts and evaluation checklists. Annotators review each task to confirm that the prompt accurately reflects the requirements of its assigned content domain and capability dimension. All verification questions are further examined to ensure they are unambiguous, visually verifiable, and consistent with the prompt specifications. We also verify that all prompts and references contain no harmful, sensitive, or personally identifiable information.

\subsection{Statistics of \ourbench}

\ourbench contains 20 evaluation tasks formed by the intersection of 5 commercial domains and 4 capability dimensions. 
Each task includes 20 curated prompts, resulting in 400 generation samples in total. 
Every prompt is paired with a checklist of 20 binary verification questions (10 easy and 10 hard), yielding 8,000 human-designed verification questions for systematic evaluation.

Figure~\ref{fig:data_statistics} illustrates the structural composition and textual characteristics of the benchmark. 
As shown in Figure~\ref{fig:data_statistics}(a), prompts for the three content-based capabilities (\textit{Text}, \textit{Layout}, and \textit{Attribute}) typically range from 200 to 1400 tokens. 
This is because these tasks require detailed descriptions of low-level visual constraints such as typography, spatial layout, element counts, and attribute specifications, which naturally lead to longer instructions. 
In contrast, prompts for \textit{Knowledge} are shorter (around 100–200 tokens), since the primary goal is to evaluate whether models can correctly incorporate domain knowledge and complete the task semantically rather than reproduce complex visual layouts. 

Figure~\ref{fig:data_statistics}(b) further demonstrates the diversity of \ourbench, where each dimension encompasses a wide spectrum of sub-attributes, from spatial properties like \textit{Position} to semantic aspects like \textit{Fact}, ensuring comprehensive coverage of fine-grained visual and logical constraints within each dimension.
Finally, the keyword frequency clouds in Figure~\ref{fig:data_statistics}(c) reveal that the most frequent terms in the verification questions align closely with the visual or semantic elements targeted by each capability dimension. 
This observation supports the validity of our checklist design and indicates that the evaluation questions effectively capture the intended capabilities of each task.

\subsection{Evaluation Criteria of \ourbench}

To evaluate complex commercial visual content, we adopt a checklist-based evaluation protocol. For each prompt in \ourbench{}, we provide a structured checklist consisting of 20 binary (Yes/No) verification questions. To automatically assess these requirements, we employ a multimodal large language model (MLLM) as the evaluation judge. Given a generated image and its corresponding checklist, the evaluator performs visual reasoning to answer each verification question. Following the benchmark design, the 20 questions are divided into \textit{easy} and \textit{hard} subsets, and we report performance on these two tracks separately.

Score for each track is computed using a penalty-based strategy: $Score=\max(0,1-\alpha N_{errors})$, where $N_{errors}$ denotes the number of incorrectly answered questions within the track and $\alpha$ is the penalty coefficient. In our benchmark we set $\alpha=0.2$, meaning that when half of the questions in a track (5 out of 10) are answered incorrectly the score becomes zero. Empirically, we observe that only a small portion of models collapse to zero scores, allowing the remaining score range to exhibit clearer performance stratification among models.

\section{Evaluation Results}

\subsection{Experimental Setup}

We evaluate \textbf{26} leading image generation models, including \textbf{10} closed-source models and \textbf{16} open-source models.
The closed-source models are: 
Nano-Banana-Pro~\cite{nanobananapro}, Nano-Banana-2.0~\cite{nanobanana2}, GPT-Image-1.5~\cite{openai2025gptimage1.5}, GPT-Image-1.0~\cite{openai_gpt_image_1}, Seedream~\cite{seedream2025seedream4,seedream45,Seedream5}, Wan2.6-T2I~\cite{wan2025}, FLUX.2-Pro~\cite{flux-2-2025} and Imagen-4~\cite{imagen4}.
The open-source models include: Z-Image~\cite{team2025zimage}, FLUX.2-dev~\cite{flux-2-2025}, Qwen-Image~\cite{wu2025qwenimage}, Emu3.5~\cite{cui2025emu35}, HunyuanImage~\cite{HunyuanImage-2.1,cao2025hunyuanimage3}, GLM-Image~\cite{glm_image}, X-Omni~\cite{geng2025xomni}, FLUX.1-schnell, FLUX.1-Krea-dev~\cite{flux1kreadev2025}, FLUX.1-dev~\cite{flux2024}, Bagel~\cite{deng2025bagel}, LongCat-Image~\cite{LongCat-Image} and SD3.5-Large~\cite{stabilityai_sd3_5}.
For all models, we use the default inference settings provided by their official APIs or documentation. The generation resolution is adaptively set to the closest supported aspect ratio of the ground-truth images to ensure structural alignment with real-world designs. Gemini-3-Flash-Preview~\cite{gemini3} is used as our automated evaluator. To reduce API overhead, all 20 checklist questions are answered in a single query. We run evaluation with a single pass per sample; despite MLLM stochasticity, repeated trials show very low variance. Gemini-3-Flash-Preview exhibits behavior similar to Gemini-3-Pro and performs better than GPT-5.2 on structured visual tasks such as chart interpretation. Detailed evaluator analysis is provided in the Appendix.

\begin{table}[t]
\caption{\textbf{Performance of Image Generative Models Across Five Commercial Content Domains.}
Closed-source and open-source models are marked with
\colorbox{closedmodel!40}{purple} and \colorbox{openmodel}{blue}, respectively. Each cell reports scores on the hard/\textcolor{easy_mode}{easy} testsets.}
\label{tab:main_per_application}
\centering
\resizebox{0.98\linewidth}{!}{%
\begin{tabular}{lcccccc}
\toprule
\textbf{Model}              & \textbf{Slides} & \textbf{Webpage} & \textbf{Poster} & \textbf{Chart} & \textbf{Sci. Figure} & \textbf{Average} \\ \hline
\closedmodel{Nano-Banana-Pro~\cite{nanobananapro}} & \scor{94.8}{82.2} & \scor{96.5}{77.5} & \scor{94.8}{76.5} & \scor{92.2}{73.0} & \scor{90.0}{74.2} & \scor{93.7}{76.7} \\
\closedmodel{Nano-Banana-2.0~\cite{nanobanana2}} & \scor{95.8}{73.8} & \scor{94.5}{71.2} & \scor{91.2}{67.5} & \scor{89.2}{60.2} & \scor{92.0}{69.5} & \scor{92.5}{68.5} \\
\closedmodel{Seedream-5.0~\cite{Seedream5}}    & \scor{80.8}{54.5} & \scor{80.8}{47.0} & \scor{77.0}{50.7} & \scor{76.2}{46.0} & \scor{81.5}{45.8} & \scor{79.2}{48.8} \\
\closedmodel{GPT-Image-1.5~\cite{openai2025gptimage1.5}}   & \scor{89.2}{40.8} & \scor{86.0}{41.0} & \scor{83.5}{42.0} & \scor{76.5}{28.2} & \scor{72.8}{27.8} & \scor{81.6}{35.9} \\
\closedmodel{Seedream-4.5~\cite{seedream45}}    & \scor{71.0}{33.8} & \scor{75.5}{36.2} & \scor{72.2}{35.5} & \scor{47.8}{18.0} & \scor{64.8}{27.3} & \scor{66.2}{30.1} \\
\closedmodel{Wan2.6-T2I~\cite{wan2025}}        & \scor{56.5}{27.1} & \scor{67.0}{25.5} & \scor{62.5}{27.5} & \scor{48.8}{17.2} & \scor{58.8}{12.5} & \scor{58.7}{21.9} \\
\closedmodel{Seedream-4.0~\cite{seedream2025seedream4}}    & \scor{67.8}{18.5} & \scor{71.8}{19.2} & \scor{65.5}{18.8} & \scor{46.0}{7.8}  & \scor{49.2}{7.2}  & \scor{60.1}{14.3} \\
\openmodel{Emu3.5~\cite{cui2025emu35}}   & \scor{44.8}{14.5} & \scor{48.8}{20.0}  & \scor{53.0}{20.3} & \scor{20.3}{4.8}  & \scor{34.0}{6.5}  & \scor{40.2}{13.2}  \\
\openmodel{HunyuanImage-3.0~\cite{cao2025hunyuanimage3}}   & \scor{47.0}{19.3} & \scor{52.0}{21.0}  & \scor{53.5}{19.8} & \scor{18.8}{2.0}  & \scor{29.0}{2.8}  & \scor{40.1}{13.0}  \\
\closedmodel{GPT-Image-1.0~\cite{openai_gpt_image_1}}   & \scor{64.8}{18.2} & \scor{63.0}{12.0} & \scor{64.8}{17.2} & \scor{31.2}{3.2}  & \scor{38.5}{5.0}  & \scor{52.4}{11.2} \\
\openmodel{HunyuanImage-2.1~\cite{HunyuanImage-2.1}}& \scor{36.2}{11.0} & \scor{40.0}{16.5} & \scor{39.0}{11.8} & \scor{8.0}{1.2} & \scor{15.0}{2.2} & \scor{27.7}{8.6} \\
\openmodel{Z-Image~\cite{team2025zimage}}         & \scor{43.8}{12.2} & \scor{48.5}{6.2}  & \scor{50.0}{12.0} & \scor{30.5}{8.8}  & \scor{46.0}{1.8}  & \scor{43.8}{8.2}  \\
\openmodel{Qwen-Image-2512~\cite{wu2025qwenimage}}   & \scor{45.0}{10.2} & \scor{48.0}{7.2}  & \scor{47.0}{11.5} & \scor{28.0}{2.2}  & \scor{37.0}{0.2}  & \scor{41.0}{6.3}  \\
\openmodel{FLUX.2-dev~\cite{flux-2-2025}}        & \scor{43.2}{5.5}  & \scor{48.8}{5.5}  & \scor{48.2}{7.2}  & \scor{33.5}{5.5}  & \scor{36.5}{0.5}  & \scor{42.0}{4.9}  \\
\openmodel{Z-Image-Turbo~\cite{team2025zimage}}   & \scor{36.5}{7.0}  & \scor{45.8}{4.5}  & \scor{40.8}{4.8}  & \scor{15.5}{0.8}  & \scor{23.2}{0.0}  & \scor{32.4}{3.4}  \\
\openmodel{Qwen-Image~\cite{wu2025qwenimage}}        & \scor{28.5}{3.5}  & \scor{27.5}{2.5}  & \scor{32.2}{5.8}  & \scor{15.8}{2.2}  & \scor{15.2}{0.0}  & \scor{23.8}{2.8}  \\
\closedmodel{FLUX.2-Pro~\cite{flux-2-2025}}        & \scor{23.0}{1.8}  & \scor{27.0}{2.0}  & \scor{26.2}{2.8}  & \scor{14.7}{1.3}  & \scor{14.3}{0.0}  & \scor{21.1}{1.6}  \\
\openmodel{GLM-Image~\cite{glm_image}}         & \scor{22.8}{1.5}  & \scor{27.3}{1.8}  & \scor{24.8}{3.8}  & \scor{0.5}{0.0}   & \scor{1.5}{0.0}   & \scor{15.3}{1.4}  \\
\closedmodel{Imagen-4~\cite{imagen4}}        & \scor{15.0}{1.5}  & \scor{14.8}{0.5}  & \scor{12.8}{1.8}  & \scor{6.8}{1.5}   & \scor{4.3}{0.2}   & \scor{10.7}{1.1}  \\
\openmodel{LongCat-Image~\cite{LongCat-Image}}         & \scor{15.0}{0.8}  & \scor{22.3}{1.8}  & \scor{21.3}{0.8}   & \scor{2.5}{0.0}   & \scor{3.8}{0.0}   & \scor{12.0}{0.7}   \\
\openmodel{X-Omni-EN~\cite{geng2025xomni}}         & \scor{9.0}{0.8}  & \scor{14.2}{0.2}  & \scor{15.8}{1.5}   & \scor{6.5}{0.0}   & \scor{1.8}{0.0}   & \scor{9.4}{0.5}   \\
\openmodel{SD3.5-Large~\cite{stabilityai_sd3_5}}        & \scor{0.5}{0.0}   & \scor{2.2}{0.2}   & \scor{2.8}{0.0}   & \scor{3.2}{1.2}   & \scor{1.5}{1.2}   & \scor{2.1}{0.5}   \\
\openmodel{Bagel~\cite{deng2025bagel}}        & \scor{4.8}{0.3}   & \scor{5.0}{0.0}  & \scor{8.5}{0.5}   & \scor{0.0}{0.0}   & \scor{0.0}{0.0}   & \scor{3.7}{0.3}   \\
\openmodel{FLUX.1-Krea-dev~\cite{flux1kreadev2025}}   & \scor{4.8}{0.0}   & \scor{8.5}{0.0}   & \scor{11.8}{0.2}  & \scor{0.8}{0.0}   & \scor{0.0}{0.0}   & \scor{5.1}{0.1}   \\
\openmodel{FLUX.1-dev~\cite{flux2024}}        & \scor{6.0}{0.0}   & \scor{11.0}{0.0}  & \scor{7.8}{0.5}   & \scor{0.2}{0.0}   & \scor{0.0}{0.0}   & \scor{5.0}{0.1}   \\
\openmodel{FLUX.1-schnell~\cite{flux2024}}    & \scor{8.5}{0.0}   & \scor{8.2}{0.0}   & \scor{8.2}{0.0}   & \scor{0.8}{0.0}   & \scor{0.0}{0.0}   & \scor{5.1}{0.0}   \\
\bottomrule
\end{tabular}
}
\end{table}

\begin{figure}[t]
    \centering
    \includegraphics[width=\linewidth]{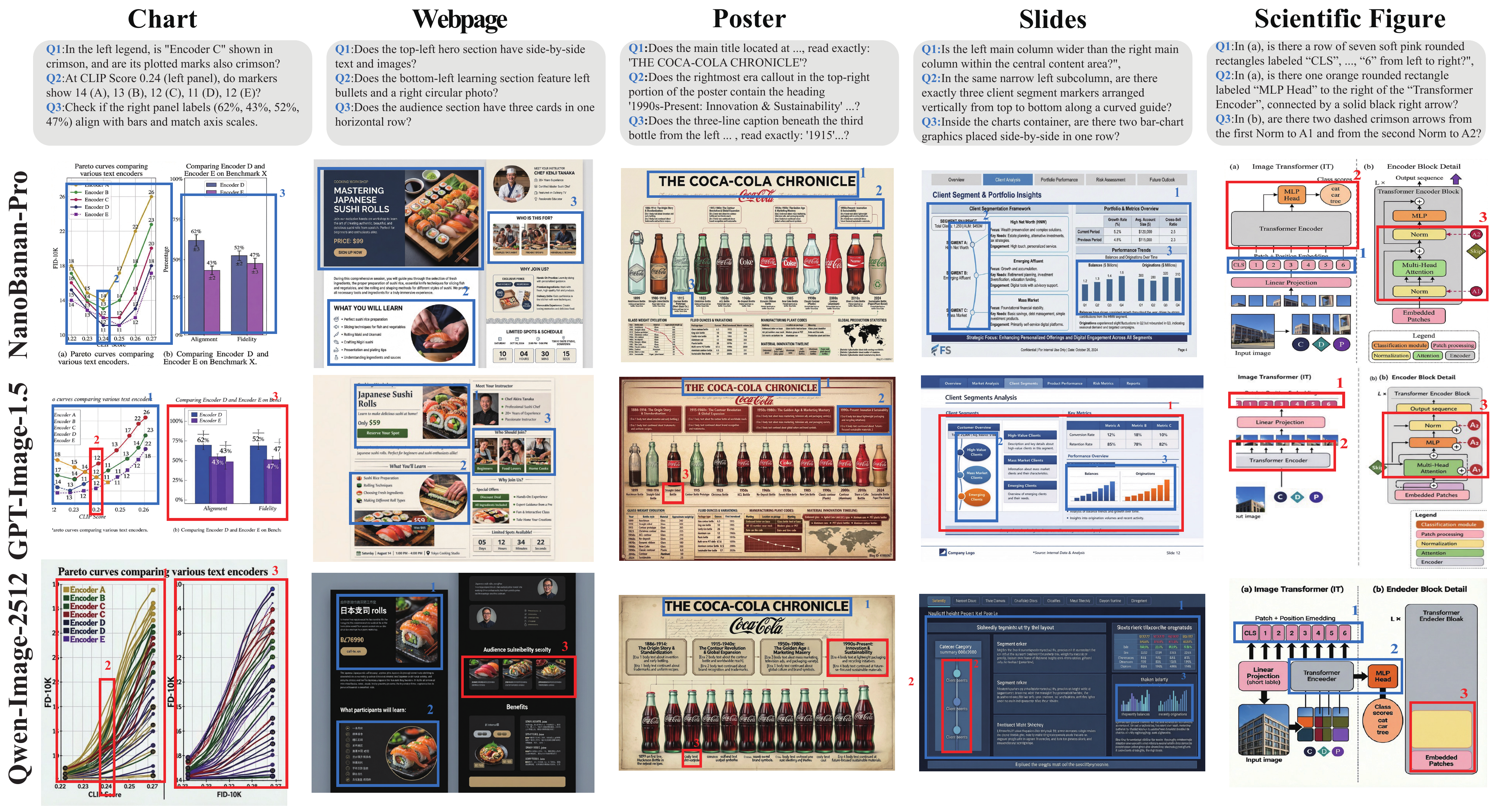}
    \caption{\textbf{Qualitative evaluation of different commercial image generation models across five content domains.} Columns represent content domains, and rows show model outputs. Evaluation questions are listed in the top row, with correct and incorrect regions highlighted in \textcolor{darkblue}{blue} and \textcolor{darkred}{red} boxes.}
    \label{fig:vis_per_applications}

\end{figure}

\subsection{Main Results}
\label{sec:main_results}

\noindent\textbf{Results for Content Domains.}
The results in Table~\ref{tab:main_per_application} highlight a pronounced sensitivity of current models to the distinct requirements of various commercial scenarios. Closed-source models, particularly Nano-Banana-Pro~\cite{nanobananapro} and Nano-Banana-2.0~\cite{nanobanana2}, consistently achieve high scores across all domains, with hard-set averages around 68–77\%. \textit{Slides}, \textit{Webpages} and \textit{Posters} are the best-performing domains, which appear well-captured by the models' training distributions. In contrast, \textit{Charts} and \textit{Scientific Figures} present greater challenges, with a noticeable drop in hard-set scores (e.g., GPT-Image-1.5~\cite{openai2025gptimage1.5} drops to 28.2–27.8\%), reflecting the difficulty of precise numerical rendering and structured diagram generation. Open-source models generally lag behind, with Z-Image~\cite{team2025zimage}, FLUX.2-dev~\cite{flux-2-2025}, and Qwen-Image-2512~\cite{wu2025qwenimage} achieving mid-range performance on \textit{Slides} and \textit{Webpages} (easy scores ~43–49\%) but near-zero results on \textit{Scientific Figures} (hard scores 0–2\%), indicating significant gaps in generating complex professional content.

As illustrated in Fig.~\ref{fig:vis_per_applications}, the qualitative gap between models is most evident in high-precision domains like \textit{Charts} and \textit{Scientific Figures}, where the ability to bind specific numerical values to spatial coordinates is a key differentiator. For example, in the Chart task (Q2), where models are asked to render \textit{``specific markers (14, 13, 12, 11, 12) at a CLIP Score of 0.24''}. Nano-Banana-Pro~\cite{nanobananapro} accurately plots and labels each individual value at its correct position. In contrast, GPT-Image-1.5~\cite{openai_gpt_image_1} exhibits a ``homogenization'' error, rendering all markers as the same incorrect value (``12''), while Qwen-Image-2512~\cite{wu2025qwenimage} fails fundamentally by omitting the numerical markers entirely.

\begin{table}[t]
\centering
\caption{\textbf{Performance of Image Generative Models Across Four Capability Dimensions.}}
\label{tab:main_per_task}
\resizebox{\linewidth}{!}{%
\begin{tabular}{lccccc}
\toprule
\textbf{Model} & \textbf{Layout} & \textbf{Attribute} & \textbf{Text} & \textbf{Knowledge} & \textbf{Average} \\
\midrule
\closedmodel{Nano-Banana-Pro~\cite{nanobananapro}} & \scor{91.2}{72.2} & \scor{92.2}{65.6} & \scor{95.0}{86.4} & \scor{96.2}{82.6} & \scor{93.7}{76.7} \\
\closedmodel{Nano-Banana-2.0~\cite{nanobanana2}} & \scor{91.0}{68.4} & \scor{91.6}{57.4} & \scor{94.6}{83.4} & \scor{93.0}{64.6} & \scor{92.5}{68.5} \\
\closedmodel{Seedream-5.0~\cite{Seedream5}}    & \scor{89.0}{67.6} & \scor{77.2}{42.4} & \scor{75.6}{43.4} & \scor{75.2}{41.8} & \scor{79.2}{48.8} \\
\closedmodel{GPT-Image-1.5~\cite{openai2025gptimage1.5}}   & \scor{84.8}{51.6} & \scor{75.2}{25.8} & \scor{82.8}{40.4} & \scor{83.6}{26.0} & \scor{81.6}{35.9} \\
\closedmodel{Seedream-4.5~\cite{seedream45}}    & \scor{71.6}{35.4} & \scor{62.8}{22.4} & \scor{72.4}{41.4} & \scor{58.2}{21.4} & \scor{66.2}{30.1} \\
\closedmodel{Wan2.6-T2I~\cite{wan2025}}        & \scor{80.6}{46.4} & \scor{60.6}{16.6} & \scor{52.6}{12.6} & \scor{41.0}{12.2} & \scor{58.7}{21.9} \\
\closedmodel{Seedream-4.0~\cite{seedream2025seedream4}}    & \scor{73.4}{27.6} & \scor{59.2}{11.4} & \scor{52.8}{11.4} & \scor{54.8}{6.8}  & \scor{60.1}{14.3} \\
\openmodel{Emu3.5~\cite{cui2025emu35}}   & \scor{63.4}{30.4} & \scor{52.6}{14.2}  & \scor{33.6}{7.0}  & \scor{11.0}{1.2}   & \scor{40.2}{13.2}  \\
\openmodel{HunyuanImage-3.0~\cite{cao2025hunyuanimage3}}   & \scor{65.0}{27.8} & \scor{53.6}{13.8}  & \scor{39.6}{10.2}  & \scor{2.0}{0.0}   & \scor{40.1}{13.0}  \\
\closedmodel{GPT-Image-1.0~\cite{openai_gpt_image_1}}   & \scor{60.2}{21.4} & \scor{48.6}{6.8}  & \scor{41.0}{8.6}  & \scor{60.0}{7.8}  & \scor{52.4}{11.2} \\
\openmodel{HunyuanImage-2.1~\cite{HunyuanImage-2.1}} & \scor{68.4}{29.0} & \scor{39.8}{5.2} & \scor{1.4}{0.0} & \scor{1.0}{0.0} & \scor{27.7}{8.6} \\
\openmodel{Z-Image~\cite{team2025zimage}}         & \scor{69.2}{26.8} & \scor{47.6}{2.6}  & \scor{45.0}{2.8}  & \scor{13.2}{0.6}  & \scor{43.8}{8.2}  \\
\openmodel{Qwen-Image-2512~\cite{wu2025qwenimage}}   & \scor{70.6}{22.2} & \scor{47.8}{1.2}  & \scor{39.2}{1.8}  & \scor{6.4}{0.0}   & \scor{41.0}{6.3}  \\
\openmodel{FLUX.2-dev~\cite{flux-2-2025}}        & \scor{67.8}{17.2} & \scor{49.2}{1.2}  & \scor{43.0}{1.0}  & \scor{8.2}{0.0}   & \scor{42.0}{4.9}  \\
\openmodel{Z-Image-Turbo~\cite{team2025zimage}}   & \scor{60.6}{11.0} & \scor{35.0}{1.2}  & \scor{29.8}{1.2}  & \scor{4.0}{0.2}   & \scor{32.4}{3.4}  \\
\openmodel{Qwen-Image~\cite{wu2025qwenimage}}        & \scor{51.2}{10.4} & \scor{22.2}{0.2}  & \scor{17.6}{0.6}  & \scor{4.4}{0.0}   & \scor{23.8}{2.8}  \\
\closedmodel{FLUX.2-Pro~\cite{flux-2-2025}}        & \scor{36.2}{6.1}  & \scor{22.9}{0.0}  & \scor{13.7}{0.0}  & \scor{11.7}{0.2}  & \scor{21.1}{1.6}  \\
\openmodel{GLM-Image~\cite{glm_image}}         & \scor{43.2}{5.4}  & \scor{13.4}{0.0}  & \scor{4.4}{0.2}   & \scor{0.4}{0.0}   & \scor{15.3}{1.4}  \\
\closedmodel{Imagen-4~\cite{imagen4}}        & \scor{26.8}{4.2}  & \scor{8.7}{0.0}   & \scor{4.0}{0.2}   & \scor{3.4}{0.0}   & \scor{10.7}{1.1}  \\
\openmodel{LongCat-Image~\cite{LongCat-Image}}        & \scor{35.8}{2.4}  & \scor{11.6}{0.2}   & \scor{4.4}{0.0}   & \scor{0.0}{0.0}   & \scor{13.0}{0.7}  \\
\openmodel{SD3.5-Large~\cite{stabilityai_sd3_5}}       & \scor{6.6}{2.2}   & \scor{0.4}{0.0}   & \scor{0.0}{0.0}   & \scor{1.2}{0.0}   & \scor{2.1}{0.5}   \\
\openmodel{X-Omni-EN~\cite{geng2025xomni}}         & \scor{22.8}{2.0}  & \scor{5.6}{0.0}   & \scor{8.0}{0.0}   & \scor{1.4}{0.0}   & \scor{9.4}{0.5}   \\
\openmodel{Bagel~\cite{deng2025bagel}}        & \scor{12.8}{0.6}  & \scor{1.6}{0.0}   & \scor{0.0}{0.0}   & \scor{0.2}{0.0}   & \scor{3.7}{0.2}   \\
\openmodel{FLUX.1-Krea-dev~\cite{flux1kreadev2025}}   & \scor{17.8}{0.2}  & \scor{2.8}{0.0}   & \scor{0.0}{0.0}   & \scor{0.0}{0.0}   & \scor{5.1}{0.1}   \\
\openmodel{FLUX.1-dev~\cite{flux2024}}        & \scor{15.8}{0.4}  & \scor{2.8}{0.0}   & \scor{0.0}{0.0}   & \scor{1.4}{0.0}   & \scor{5.0}{0.1}   \\
\openmodel{FLUX.1-schnell~\cite{flux2024}}    & \scor{16.8}{0.0}  & \scor{2.6}{0.0}   & \scor{0.0}{0.0}   & \scor{1.2}{0.0}   & \scor{5.1}{0.0}   \\
\bottomrule
\end{tabular}%
}
\end{table}

\begin{figure}[t]
    \centering
    \includegraphics[width=\linewidth]{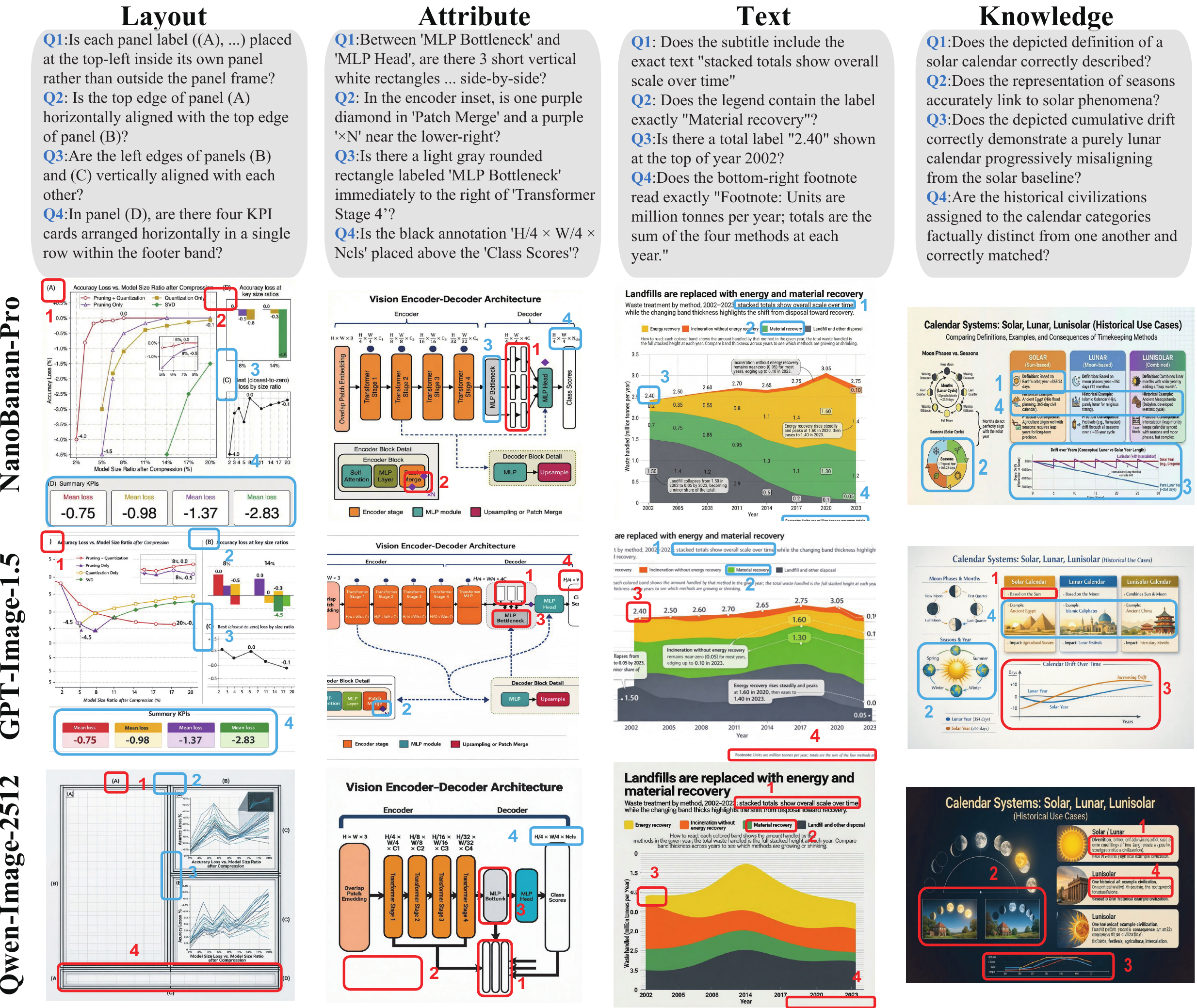}
    \caption{\textbf{Qualitative evaluation of different commercial image generation models across four capability dimensions.}}
    \label{fig:vis_per_task}

\end{figure}

\noindent\textbf{Results for Capability Dimensions.}
Table~\ref{tab:main_per_task} reports model performance across the four capability dimensions: \textit{Layout Control}, \textit{Attribute Binding}, \textit{Text Rendering}, and \textit{Knowledge-based Reasoning}. Closed-source models generally perform well on \textit{Text Rendering} and \textit{Knowledge-based Reasoning}. For instance, Nano-Banana-Pro~\cite{nanobananapro} achieves \score{95.0}{86.4} on \textit{Text Rendering} and \score{96.2}{82.6} on \textit{Knowledge-based Reasoning}, indicating strong textual fidelity and domain reasoning.
In contrast, \textit{Layout Control} and \textit{Attribute Binding} remain more challenging, especially on the hard subsets (e.g., 72.2 for Layout and 65.6 for Attribute), revealing persistent limitations in fine-grained spatial reasoning and precise property control. This difficulty is more evident for open-source models. While Z-Image~\cite{team2025zimage} and FLUX.2-dev~\cite{flux-2-2025} retain basic layout capability (easy scores around 67.8–69.2\%), their performance on \textit{Knowledge-based Reasoning} and \textit{Attribute Binding} drops sharply, often approaching zero on hard questions. These results suggest that although modern models can generate legible text and perform basic semantic reasoning, producing structurally precise and attribute-consistent designs remains a key challenge for commercial visual generation.

As illustrated in Fig.~\ref{fig:vis_per_task}, qualitative examples further highlight these limitations. Even top-tier models struggle with complex layout constraints and precise attribute control. Although Nano-Banana-Pro~\cite{nanobananapro} produces more coherent structures, it still fails in highly specific spatial instructions (red boxes). For example, the Layout task (Q1) requires \textit{``placing panel labels (e.g., `(A)') strictly inside the top-left corner of each panel frame''}, yet all three models fail to maintain the required boundary logic.

\noindent\textbf{Human Evaluation.}
We conducted a human evaluation on a subset of benchmark tasks. From the benchmark pool, 2,000 questions were randomly sampled (400 per model) covering Nano-Banana-Pro~\cite{nanobananapro}, GPT-Image-1.5~\cite{openai2025gptimage1.5}, Seedream-5.0~\cite{Seedream5}, Qwen-Image-2512~\cite{wu2025qwenimage}, and Z-Image~\cite{team2025zimage}. For each participant, two questions per model were randomly selected (10 questions total). Two additional fixed, intentionally simple reference questions were included to detect unreliable responses, resulting in 12 questions per participant. The study involved 59 participants with prior experience in visual design or data interpretation. After removing 9 responses that failed the quality checks, the observed agreement was $p_o = 90.88\%$ with Cohen's $\kappa = 0.7692$, indicating strong consistency between the automated evaluator and human judgments.

\subsection{Main Findings}

\textit{1) Stylistic Generation Does Not Imply Precise Composition.} 
Although current generative models capture the high-level stylistic patterns of commercial documents well, they still lack deterministic control required for precise composition. This limitation appears in both spatial geometry and attribute binding. As in Fig.~\ref{fig:vis_per_applications}, even top-tier models such as Nano-Banana-Pro~\cite{nanobananapro} struggle with tasks requiring precise element localization (e.g., Layout Q1\&Q2) and accurate attribute counting (e.g., Attribute Q1\&Q2), indicating that many models approximate layouts stylistically rather than enforcing strict structural constraints.

\textit{2) Text and Knowledge Capabilities Are Highly Polarized Across Models.} 
Top-tier commercial APIs such as Nano-Banana-Pro~\cite{nanobananapro} (\score{95.0}{86.4} in Text and \score{96.2}{82.6} in Knowledge) achieve strong results on language-intensive tasks, likely benefiting from integration with multimodal foundation models. For example, Gemini-Image-Pro documentation states that Nano Banana Pro is built on Gemini 3 Pro and leverages its reasoning and world knowledge for visual generation. In contrast, most models perform dramatically worse: as shown in Table~\ref{tab:main_per_task}, 21 out of 26 evaluated models score below 12.6 in both Text and Knowledge, with several approaching zero. Notably, all open-source models fall into this regime, revealing both the difficulty of accurate text and knowledge grounding and a capability gap between open-source models and closed-source commercial APIs.

\begin{wraptable}{r}{0.55\columnwidth}
\centering
\scriptsize
\setlength{\tabcolsep}{3pt}
\caption{\textbf{Cross-benchmark comparison.}}
\label{tab:compare_benchmark}
\resizebox{0.55\columnwidth}{!}{
\begin{tabular}{lccc}
\toprule
Model & GenEval~\cite{ghosh2023geneval} & OneIG-EN~\cite{chang2025oneig} & \ourbench \\
\midrule
HunyuanImage-3.0~\cite{cao2025hunyuanimage3} & 0.72 & - & \scor{40.1}{13.0} \\
GPT-Image-1.0~\cite{openai_gpt_image_1} & 0.84 & 0.533 & \scor{52.4}{11.2} \\
Z-Image~\cite{team2025zimage} & 0.84 & 0.546 & \scor{43.8}{8.2} \\
Qwen-Image~\cite{wu2025qwenimage} & 0.87 & 0.539 & \scor{23.8}{2.8} \\
LongCat-Image~\cite{LongCat-Image} & 0.87 & - & \scor{12.0}{0.7} \\
\bottomrule
\end{tabular}}
\end{wraptable}

\textit{3) Natural Image Competence Does Not Transfer to Commercial Documents.} 
Models trained on natural images show clear limitations when applied to commercial visual documents. As shown in Table~\ref{tab:compare_benchmark}, GPT-Image-1.0 and Qwen-Image achieve strong scores on GenEval~\cite{ghosh2023geneval} (0.84 and 0.87 respectively), yet perform differently on \ourbench{} (\score{52.4}{11.2} vs.\ \score{23.8}{2.8}). This discrepancy indicates that strong performance on natural-image benchmarks does not directly translate to professional design scenarios, which require dense text rendering, structured layouts, and multiple compositional constraints. Although OneIG-Bench~\cite{chang2025oneig} includes prompts related to slides or posters, its score differences remain relatively small and fail to reveal the substantial capability gaps exposed by \ourbench{}.

\section{Conclusion}
In this work, we introduced \ourbench{}, a systematic benchmark for evaluating image generation models on commercial visual content creation. By covering five representative document types and four capability dimensions, \ourbench{} provides a structured evaluation framework with diverse tasks, carefully curated prompts, and rigorous checklist-based verification. Through large-scale benchmarking of both state-of-the-art commercial APIs and open-source models, our results reveal substantial gaps between current generative systems and the requirements of real-world commercial design tasks, particularly in structured layout composition, precise attribute control, and complex multi-constraint generation. We hope \ourbench{} can serve as a standardized benchmark to facilitate future research on practical visual content generation and promote the development of generative models better aligned with real-world design applications.

%
%
\bibliographystyle{splncs04}
\bibliography{main}

\newpage
\appendix
\section*{Supplemental Materials for \ourbench{}}

In these supplemental materials, we provide additional experimental data, model performance rankings, comprehensive evaluator reliability studies and extended qualitative analyses. The contents are organized as follows:

\begin{itemize}
    \item \textbf{Sec.~\ref{sec:examples}:} Visual examples of \ourbench samples, illustrating the diverse prompts and corresponding multi-dimensional checklist questions.
    \item \textbf{Sec.~\ref{sec:leaderboard_vis}:} Extensive model performance rankings across all five content domains and four capability dimensions.
    \item \textbf{Sec.~\ref{sec:evaluator_reliability}:} In-depth analysis of MLLM-based evaluation, including stability tests and alignment with human judgment.
\end{itemize}

\section{Visual Examples of \ourbench}
\label{sec:examples}

\begin{figure}[ht]
    \centering
    \begin{minipage}{\linewidth}
        \begin{promptbox}
            \textbf{Prompt:} Generate a publication-style figure (panels A--F) titled ``Probability (Harder): Bayes' Theorem and the Base Rate Fallacy''.
Problem statement (Panel A):
- A disease has a prevalence of 0.1\% in the general population. A diagnostic screening test has a sensitivity of 99\% and a false positive rate of 5\%. If a random person tests positive, what is the exact probability they actually have the disease?
Panel requirements:
- Panel A: A complete probability tree diagram showing Disease (D) and No Disease ($\sim$D) branches, followed by Test Positive (+) and Test Negative (-) sub-branches, with all edge weights labeled.
- Panel B: The formal mathematical statement of Bayes' Theorem for P(D | +).
- Panel C: Calculation of the total marginal probability of a positive test P(+), explicitly separating the True Positive and False Positive components.
- Panel D: Numeric substitution into the Bayes' formula.
- Panel E: The final calculated probability expressed as a percentage.
- Panel F: An array diagram or pie chart visually demonstrating the 'Base Rate Fallacy' (showing physically why the vast majority of positive results in this scenario are false positives).
Content requirements:
- The probability logic and arithmetic must be flawlessly executed. No placeholders; strict journal style.
        \end{promptbox}
        
        \includegraphics[width=\linewidth]{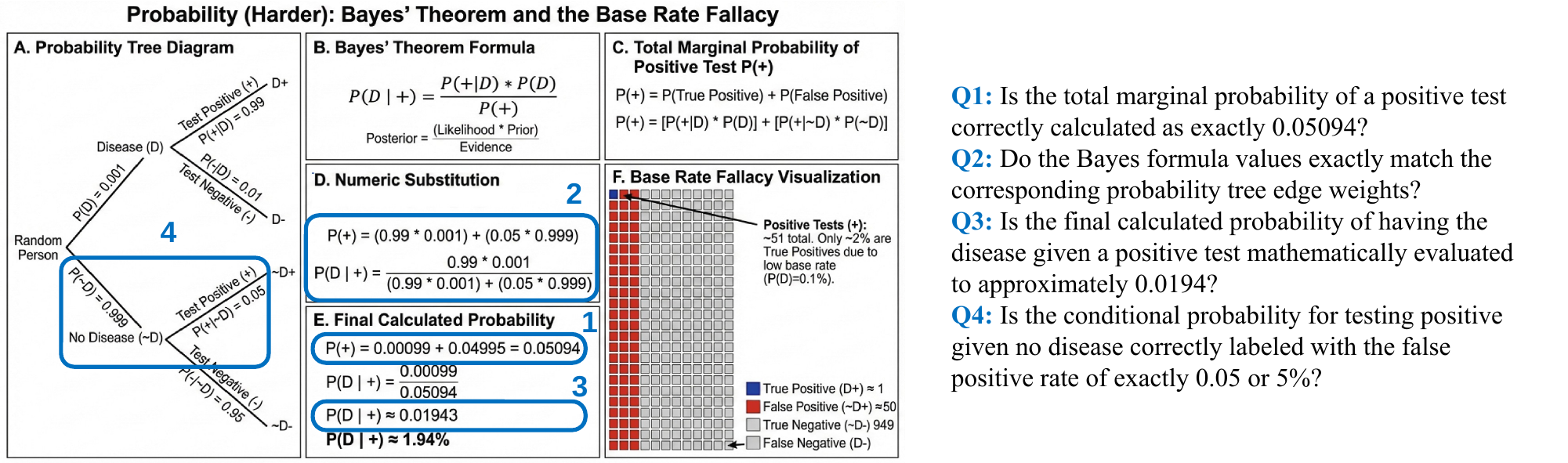}
        \caption{\textbf{Example Task: Knowledge-based Reasoning.} Assessment of a model's ability to incorporate domain knowledge into visual generation.}
        \label{fig:eg_knowledge}
    \end{minipage}
\end{figure}

\begin{figure}[ht]
    \centering
    \begin{minipage}{\linewidth}
        \begin{promptbox}
            \textbf{Prompt:} Create a flat 2D infographic on a solid light cream background... near the top, draw a wide horizontal rounded rectangle... filled with a solid burnt orange color... center the white all-caps title ``BUSINESS NAME GOES HERE''; directly below... add a second wide horizontal rounded rectangle... filled with solid dark teal... center within it all-caps white text ``CUSTOMER SERVICE PILLARS''; beneath this... arrange a row of exactly seven rounded squares, each a bright warm pink tile... center in each a black line icon: 1. speech bubbles, 2. trophy, 3. silhouettes with heart, 4. human head, 5. light bulb with gear, 6. handshake, and 7. eye; directly below these... place a second row of seven vertical rounded rectangles... filled with solid soft tangerine orange; in the first rectangle, write heading ``COMMUNICATION'' and body text ``Clear and professional verbal or written communication...''; in the remaining six... center the heading ``PILLAR GOES HERE'' and placeholder text ``Supporting details and examples go here.''; ensure all text uses a modern geometric sans-serif... forming a precise seven-column grid with equal horizontal spacing.
        \end{promptbox}
        
        \includegraphics[width=\linewidth]{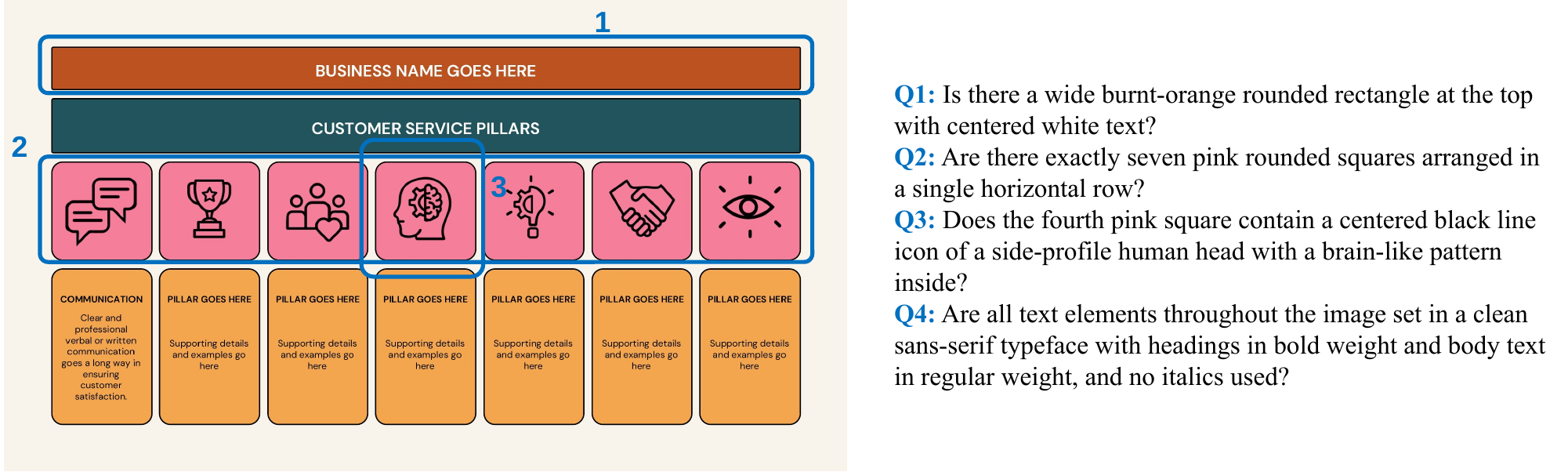}
        \caption{\textbf{Example Task: Attribute Binding.} Evaluation of precise associations between specific visual elements and their described properties.}
        \label{fig:eg_attribute}
    \end{minipage}
\end{figure}

\begin{figure}[ht]
    \centering
    \begin{minipage}{\linewidth}
        \begin{promptbox}
            \textbf{Prompt:} Create a clean two-panel scientific figure... on a white background... Panel (a) is on the left...caption '(a) Model architecture'...panel (b) is on the right...caption '(b) Detection example'... In panel (a)... four large rounded-rectangle regions... labeled: 'backbone', 'encoder', 'decoder', and 'prediction heads'... Inside 'backbone', place... overlapping light-gray rectangles labeled 'input image'... draw one black solid arrow... to a right-pointing trapezoid filled orange labeled 'CNN'... draw one black solid arrow... to a tall thin vertical rectangle labeled 'image features'... Below, draw... rounded rectangle with a teal gradient fill labeled 'positional encoding'... connect it to a small circle... containing a plus sign '+' using a black solid arrow, and also connect 'image features' to the same plus-circle... From the plus circle, draw a short dotted line to a single horizontal row... of six small light-gray squares... from the right side... draw one black solid arrow into a large rectangle labeled 'transformer encoder' located inside the 'encoder' region... To the right inside 'decoder', draw one large rectangle labeled 'transformer decoder' and connect the encoder box to the decoder box... Under the decoder box, draw exactly six small square icons in one row for queries, colored... In the top-left corner... add a tiny legend... 'bird 1' and... 'bird 2'.
        \end{promptbox}
        
        \includegraphics[width=\linewidth]{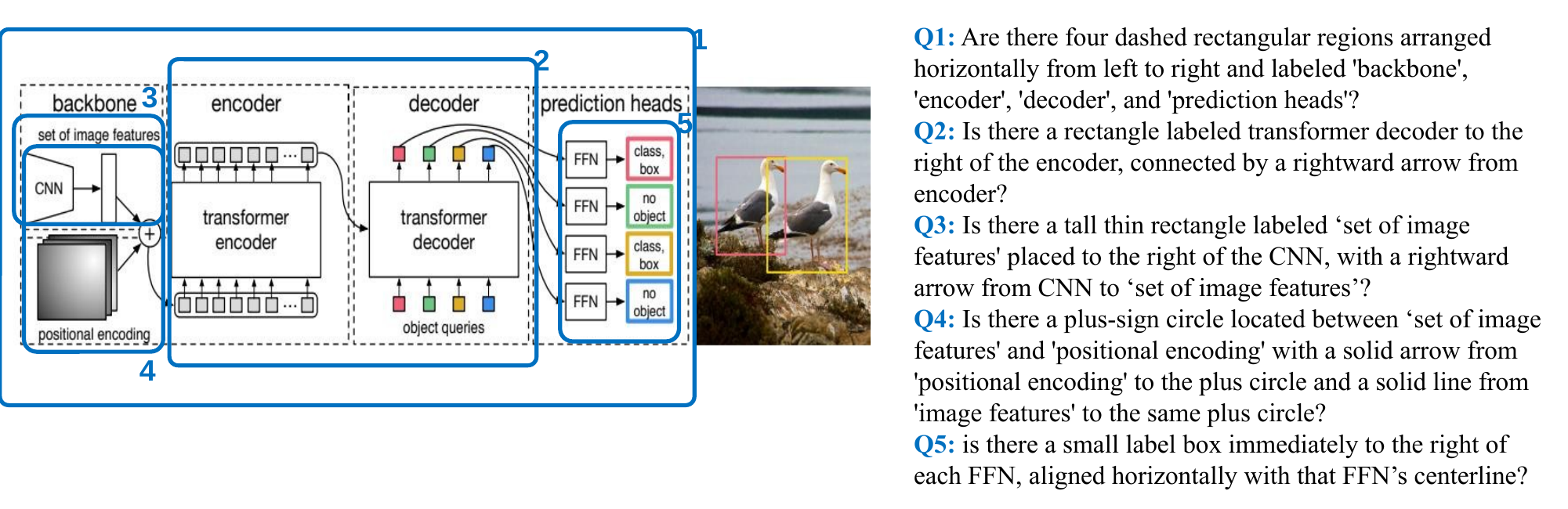}
        \caption{\textbf{Example Task: Layout Control.} Measurement of spatial arrangement and structural organization within professional design contexts.}
        \label{fig:eg_layout}
    \end{minipage}
\end{figure}

\begin{figure}[ht]
    \centering
    \begin{minipage}{\linewidth}
        \begin{promptbox}
            \textbf{Prompt:} Create an infographic-style image laid out in two main vertical columns... tall left panel slightly below the top, place a large three-line title... reading: ``6 REASONS'', ``TO GET UP'', ``EARLY''... At the upper-right of the entire image, create the first reason section... heading reads ``More time for yourself''; directly underneath that heading place one continuous paragraph that reads exactly: ``You might be able to get some much-needed (and desperately wanted) time for yourself...''. Below this, in the middle-right panel, add the second reason section... text with a single-line heading that reads ``More time to exercise''; directly beneath this heading include one paragraph of body text that reads exactly: ``Getting up early could help you find the time to work out...''. To the left of this central right panel... insert the third reason section panel... heading at the top... must read ``Less time in traffic''... and the paragraph directly below it must read exactly: ``Early mornings can help you beat the usual traffic...''. Under the middle-right panel... create the fourth reason section... text block with a heading that reads on one line ``More time to get things done'' and with a paragraph immediately below that reads exactly: ``Have you ever wished there were more hours in the day?...''. On the lower-left of the entire image... insert the fifth reason panel... text block whose heading reads ``Helps you concentrate''... with a paragraph directly below that reads exactly: ``We tend to feel groggy and disoriented when
        \end{promptbox}
        
        \includegraphics[width=\linewidth]{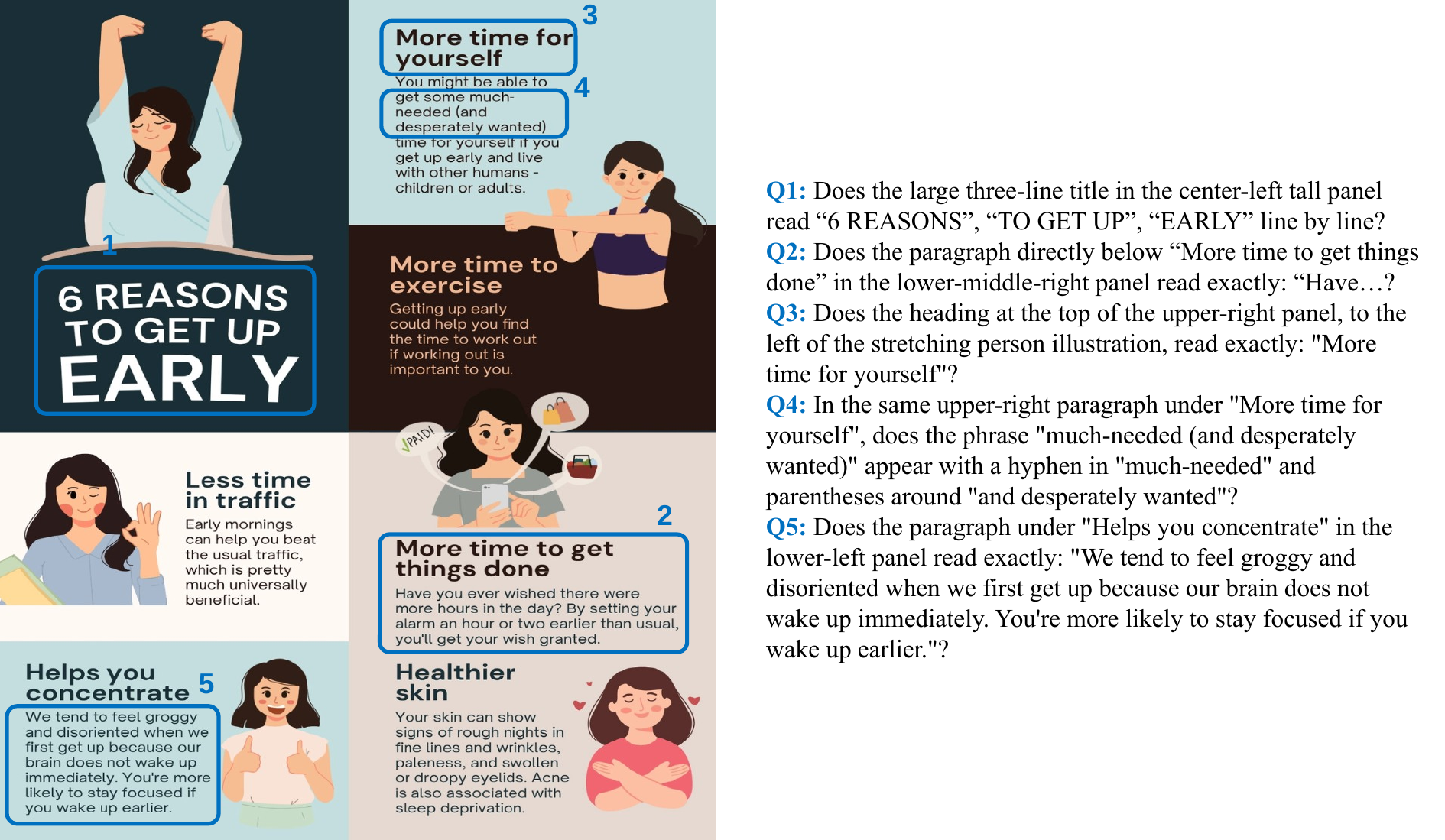}
        \caption{\textbf{Example Task: Text Rendering.} Focus on the model's capacity for legible, accurate, and typographically coherent text generation.}
        \label{fig:eg_text}
    \end{minipage}
\end{figure}

We present representative samples from \ourbench{} in Figures~\ref{fig:eg_knowledge}--\ref{fig:eg_text}. Each sample consists of a detailed prompt and its corresponding human-verified verification questions. These examples span the four key capability dimensions: \textit{Knowledge-based Reasoning} (Fig.~\ref{fig:eg_knowledge}), \textit{Attribute Binding} (Fig.~\ref{fig:eg_attribute}), \textit{Layout Control} (Fig.~\ref{fig:eg_layout}), and \textit{Text Rendering} (Fig.~\ref{fig:eg_text}).

\section{Model Performance Comparison Across Domains}
\label{sec:leaderboard_vis}

\begin{figure}[ht]
    \centering
    \includegraphics[width=0.95\linewidth]{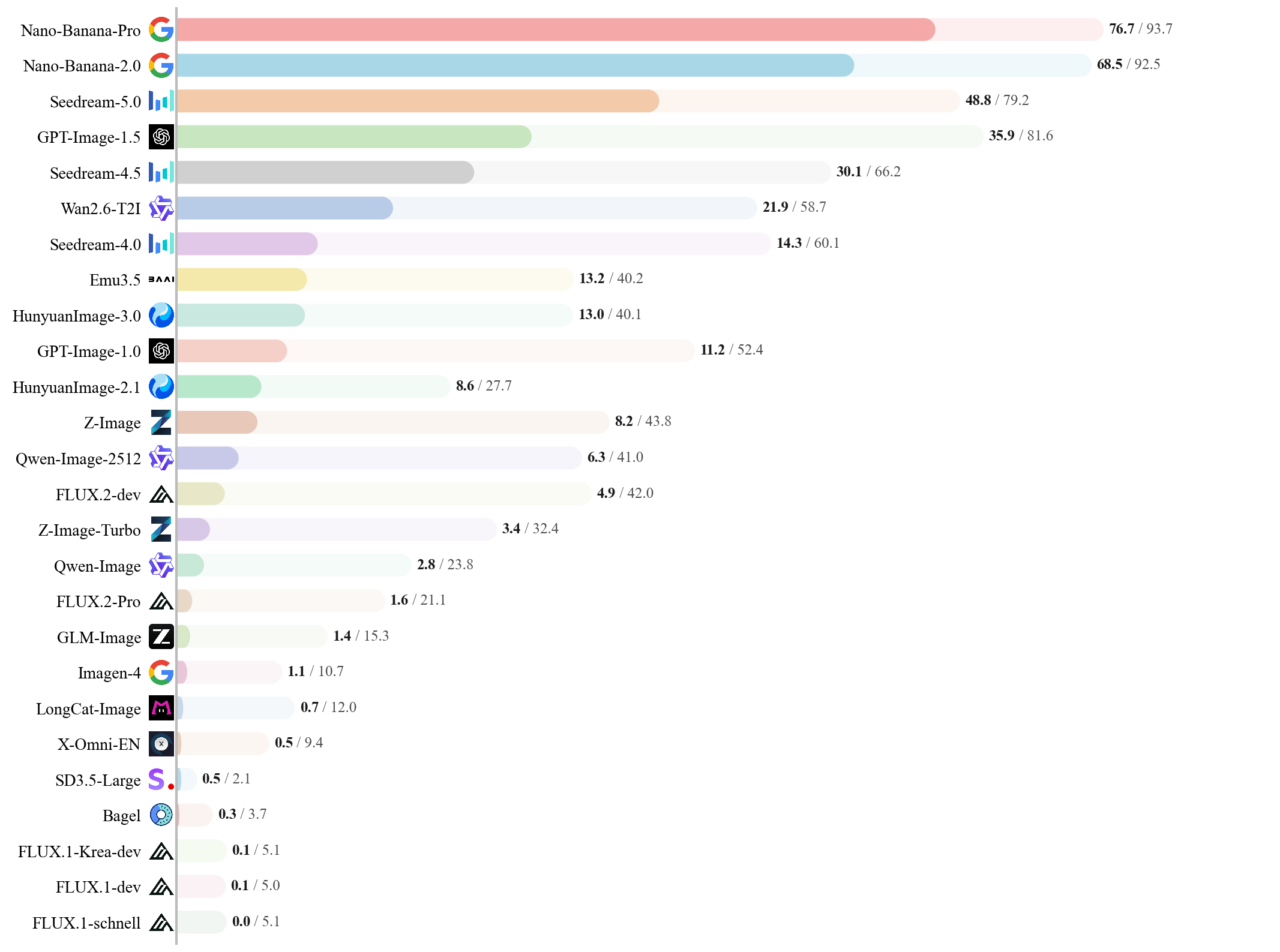}
    \caption{\textbf{Overall Ranking.}}
    \label{fig:rank_overall}
\end{figure}

We provide a comprehensive performance ladder visualization to detail the landscape of current model capabilities. Figure~\ref{fig:rank_overall} shows the overall ranking, Figures~\ref{fig:rank_webpage}-\ref{fig:rank_sci} present performance rankings across five professional content domains, and Figures~\ref{fig:rank_layout_cap}-\ref{fig:rank_know_cap} show the aggregated results across four capability dimensions. These rankings establish a clear hierarchy of current \textit{state-of-the-art} models and highlight the relative difficulty of realistic commercial design tasks.

\subsection{Comparison across content domains}
\begin{figure}[ht]
    \centering
    \includegraphics[width=0.95\linewidth]{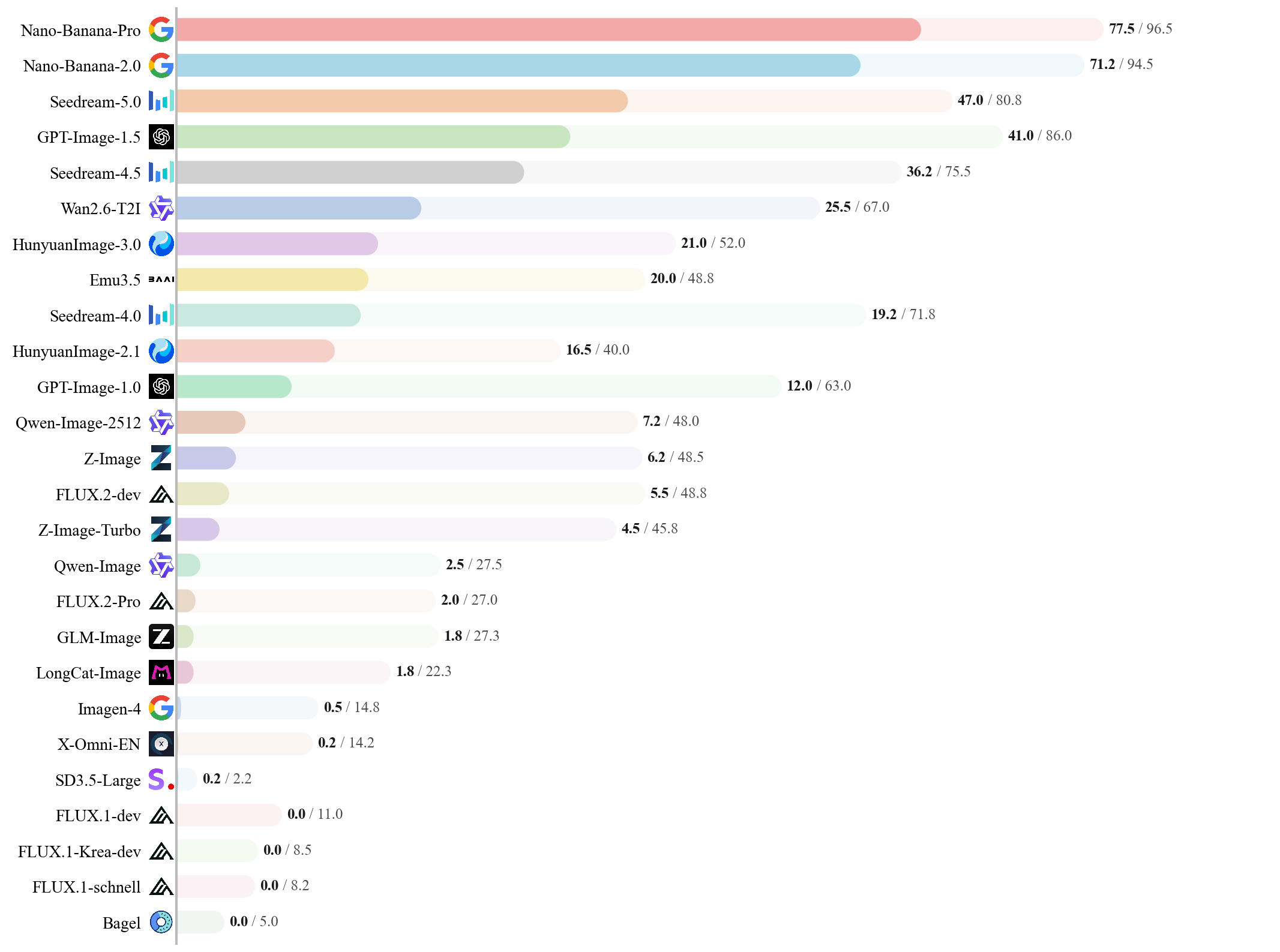}
    \caption{\textbf{Performance Ranking: Webpage.}}
    \label{fig:rank_webpage}
\end{figure}

\begin{figure}[ht]
    \centering
    \includegraphics[width=0.95\linewidth]{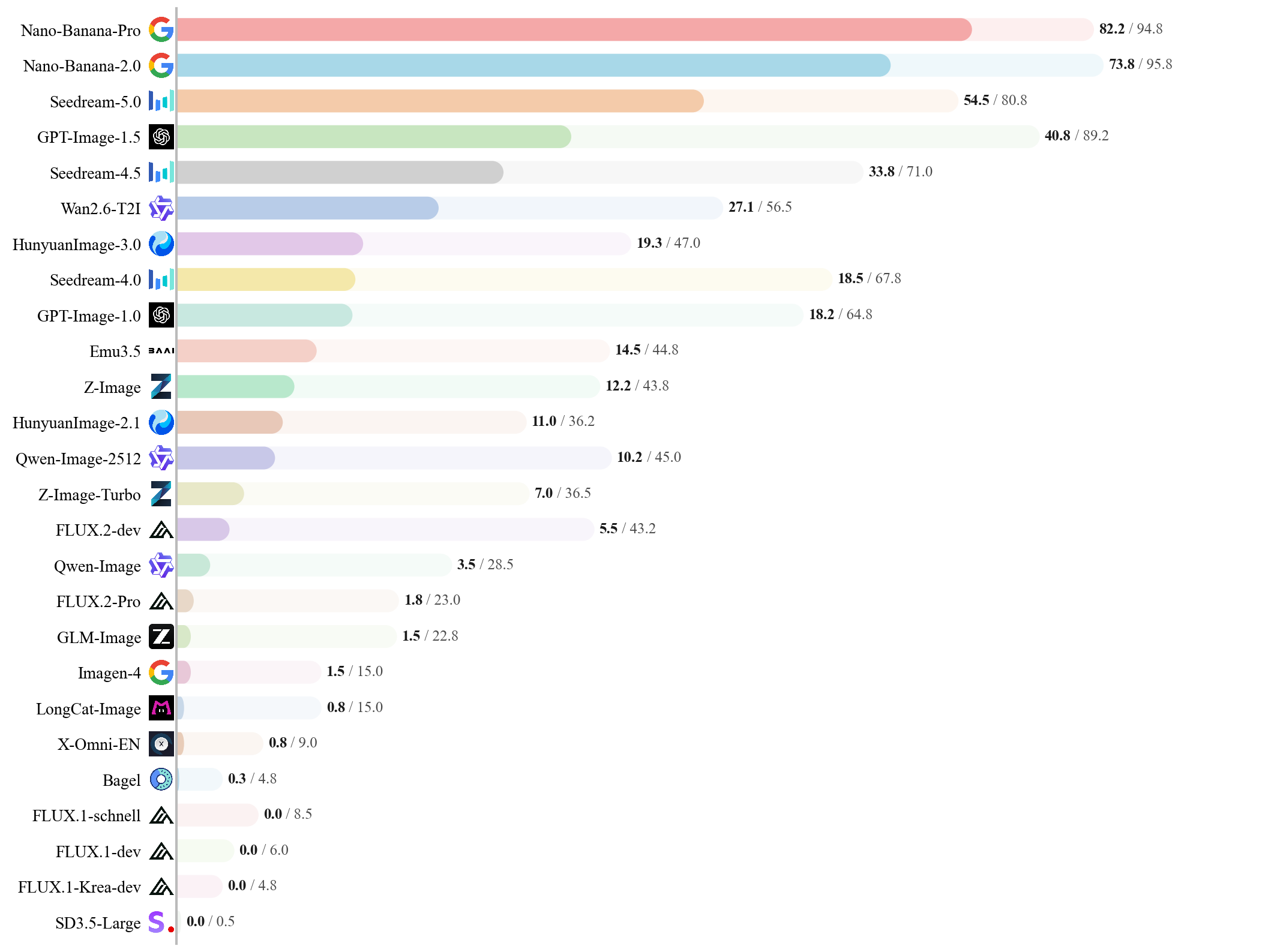}
    \caption{\textbf{Performance Ranking: Slides.}}
    \label{fig:rank_slides}
\end{figure}

\begin{figure}[ht]
    \centering
    \includegraphics[width=0.95\linewidth]{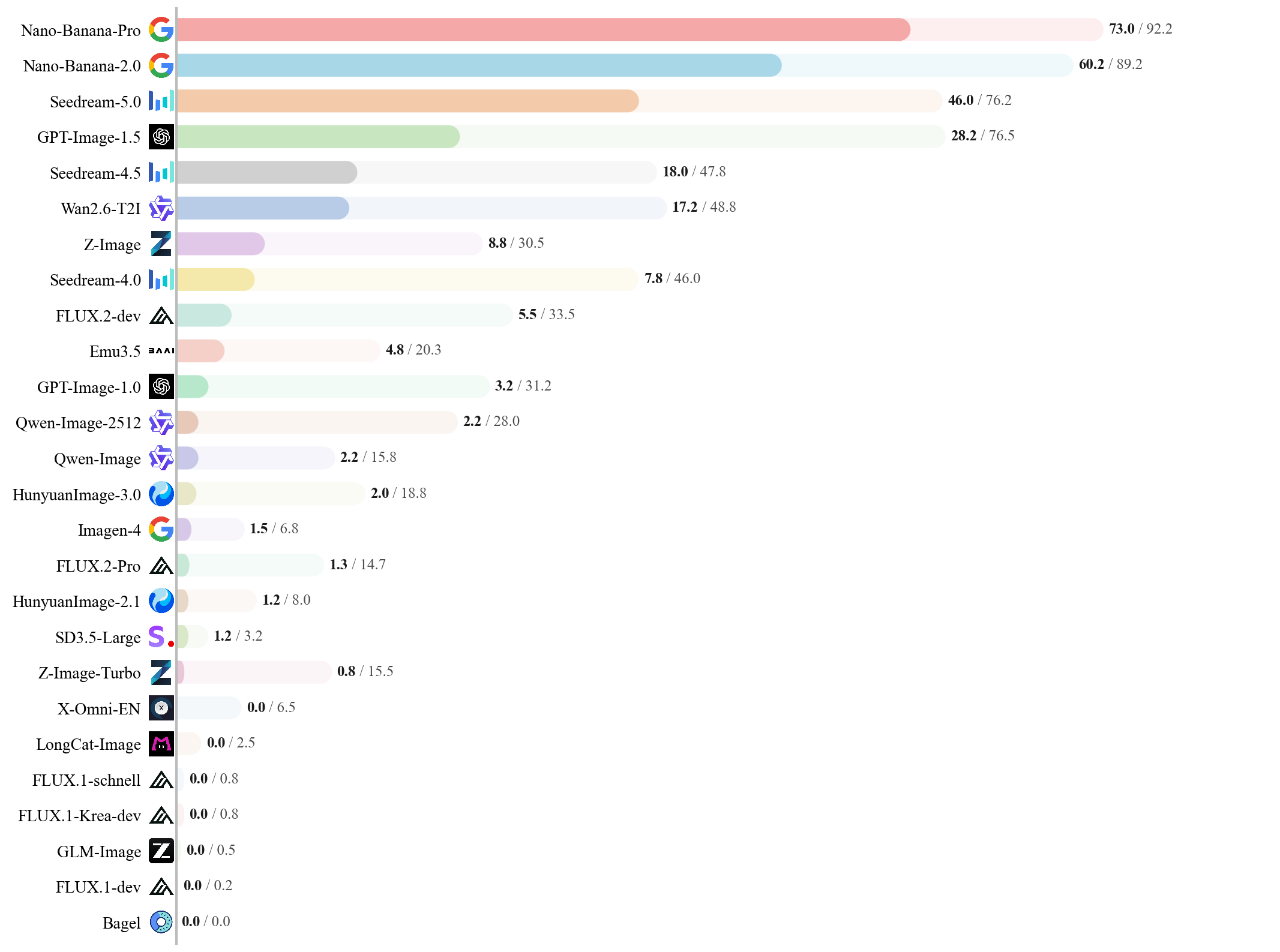}
    \caption{\textbf{Performance Ranking: Chart.}}
    \label{fig:rank_chart}
\end{figure}

\begin{figure}[ht]
    \centering
    \includegraphics[width=0.95\linewidth]{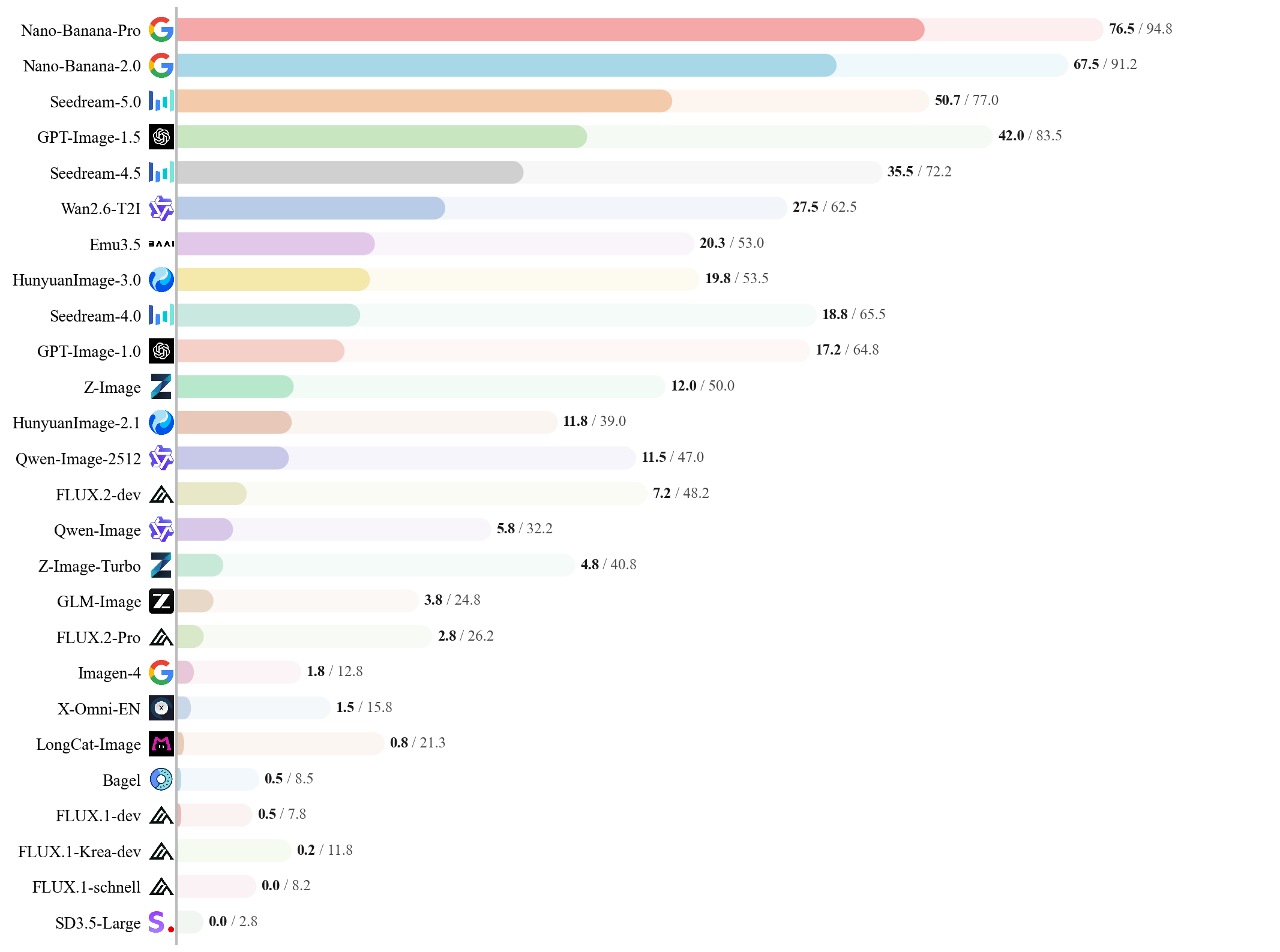}
    \caption{\textbf{Performance Ranking: Poster.}}
    \label{fig:rank_poster}
\end{figure}

\begin{figure}[ht]
    \centering
    \includegraphics[width=0.95\linewidth]{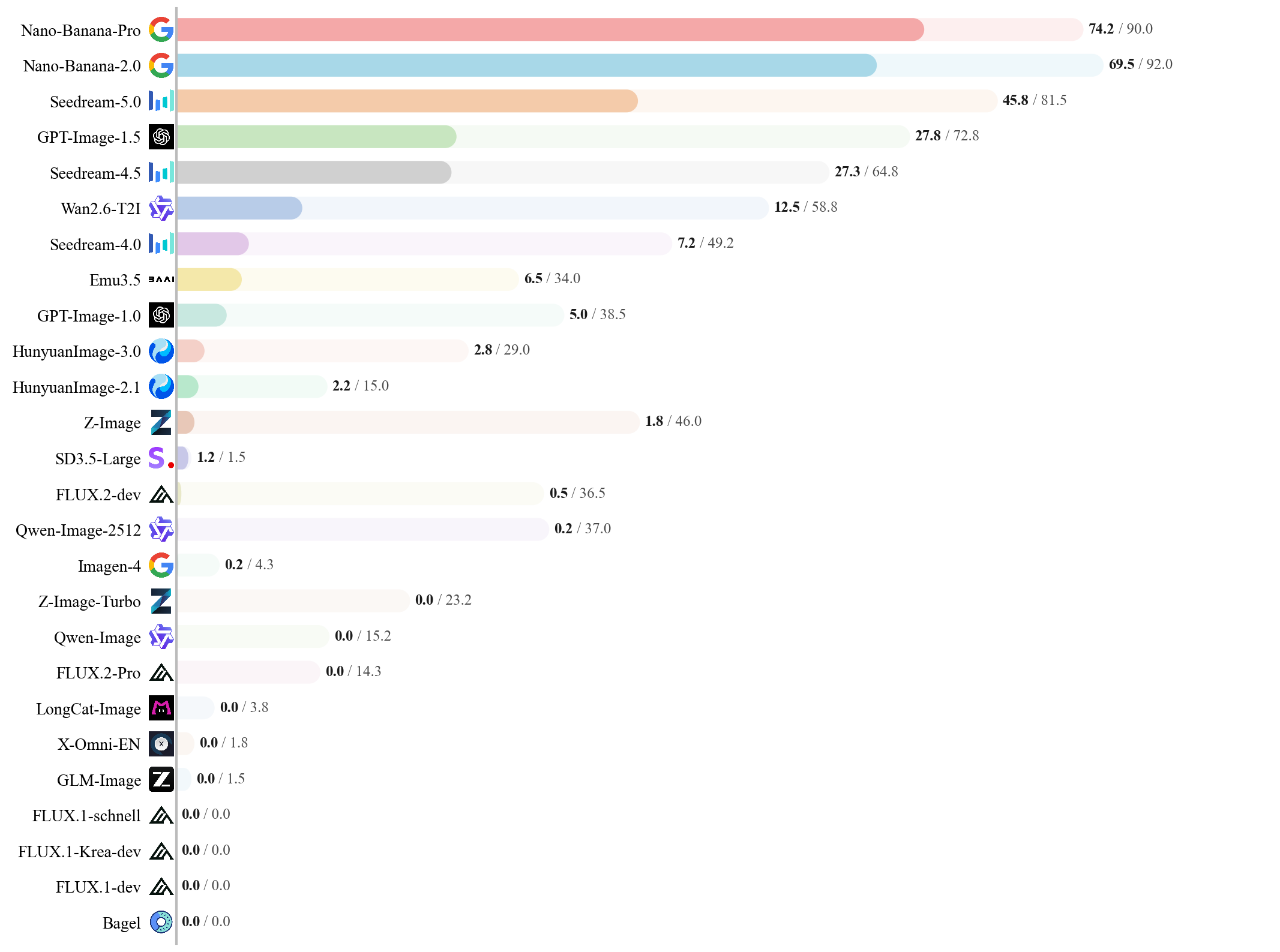}
    \caption{\textbf{Performance Ranking: Scientific Figure.}}
    \label{fig:rank_sci}
\end{figure}

We evaluate the performance across specific application scenarios, ranking models based on their ability to handle domain-specific constraints in Webpage, Slides, Chart, Poster, and Scientific Figure as shown in Figures~\ref{fig:rank_webpage}, \ref{fig:rank_slides}, \ref{fig:rank_chart}, \ref{fig:rank_poster} and \ref{fig:rank_sci}, respectively. For each model presented in the rankings, the bar with deeper color represents the performance on the ``hard'' subset, while lighter color denotes the ``easy'' subset score.

\subsection{Comparison across capability dimensions}
\begin{figure}[ht]
    \centering
    \includegraphics[width=0.95\linewidth]{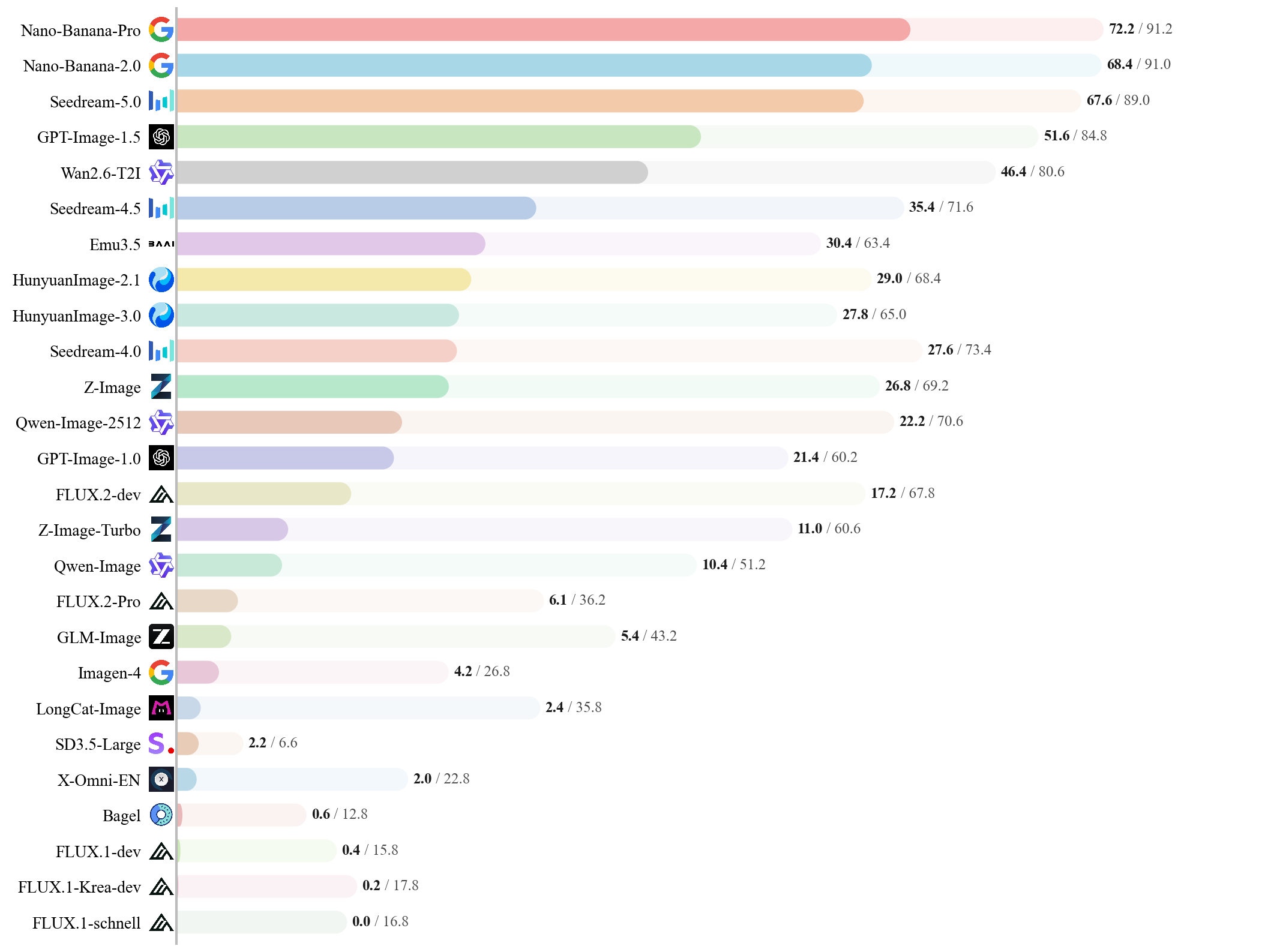}
    \caption{\textbf{Capability Ladder: Layout Control.}}
    \label{fig:rank_layout_cap}
\end{figure}

\begin{figure}[ht]
    \centering
    \includegraphics[width=0.95\linewidth]{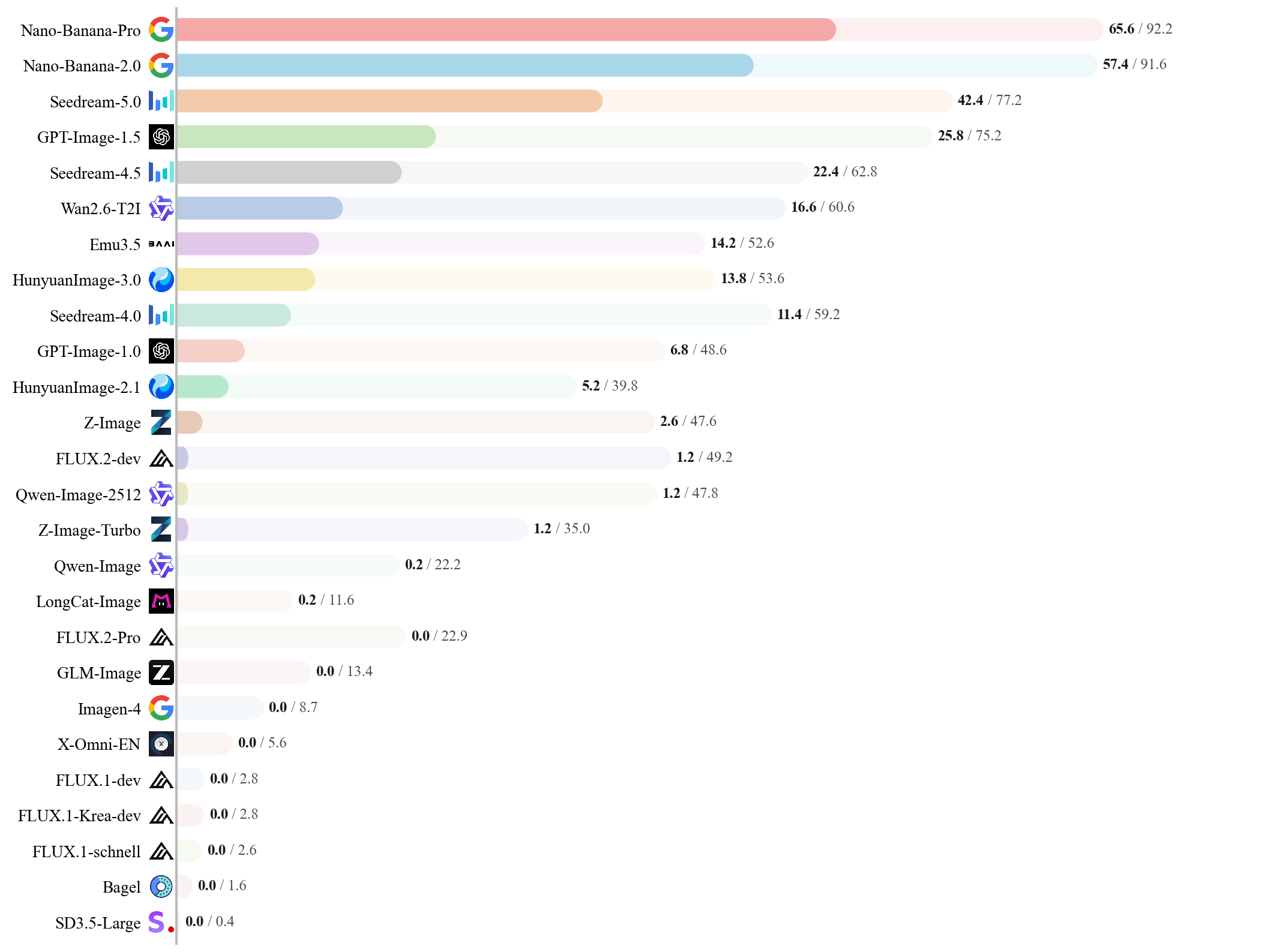}
    \caption{\textbf{Capability Ladder: Attribute Binding.}}
    \label{fig:rank_attr_cap}
\end{figure}

\begin{figure}[ht]
    \centering
    \includegraphics[width=0.95\linewidth]{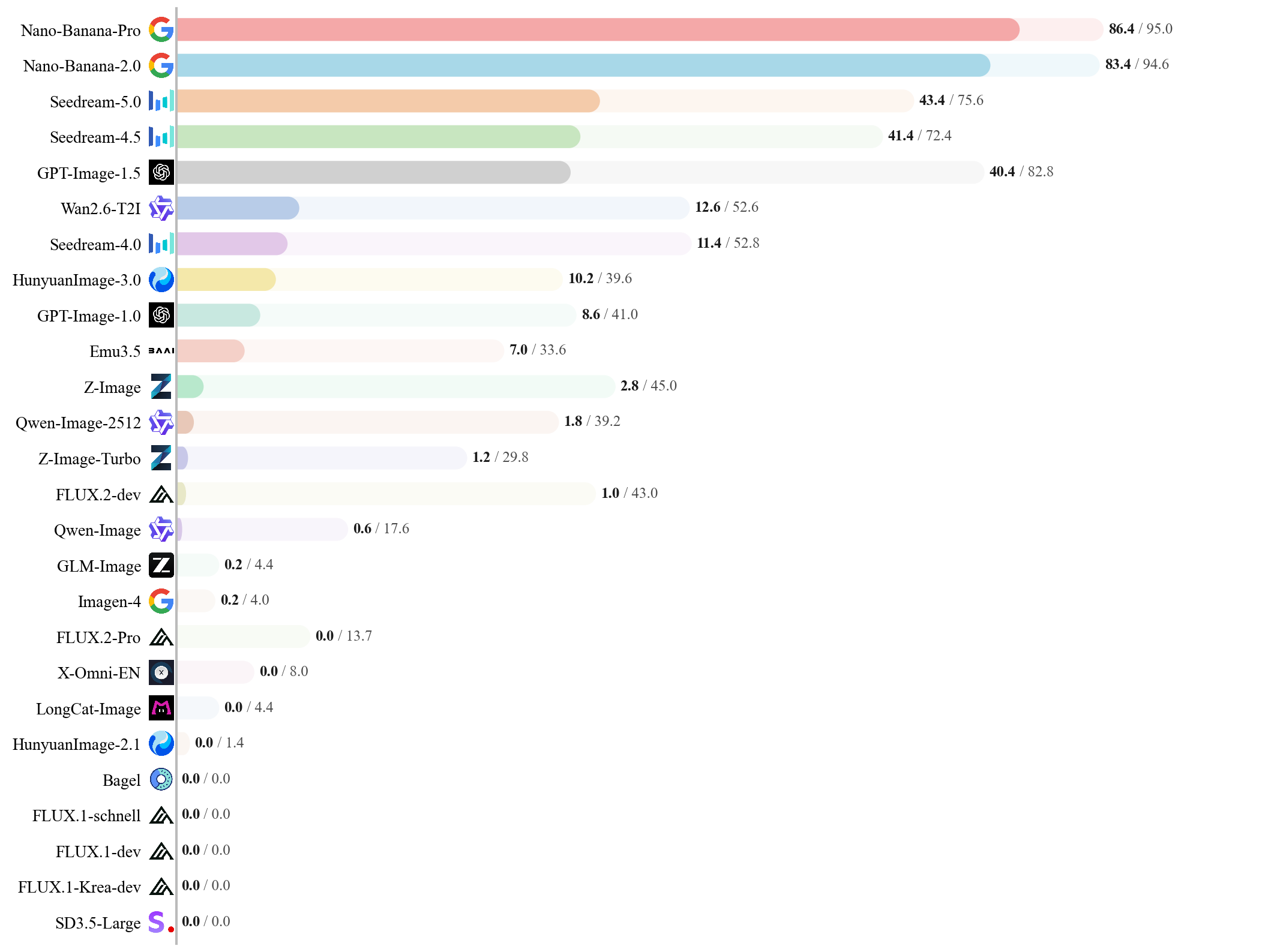}
    \caption{\textbf{Capability Ladder: Text Rendering.}}
    \label{fig:rank_text_cap}
\end{figure}

\begin{figure}[ht]
    \centering
    \includegraphics[width=0.95\linewidth]{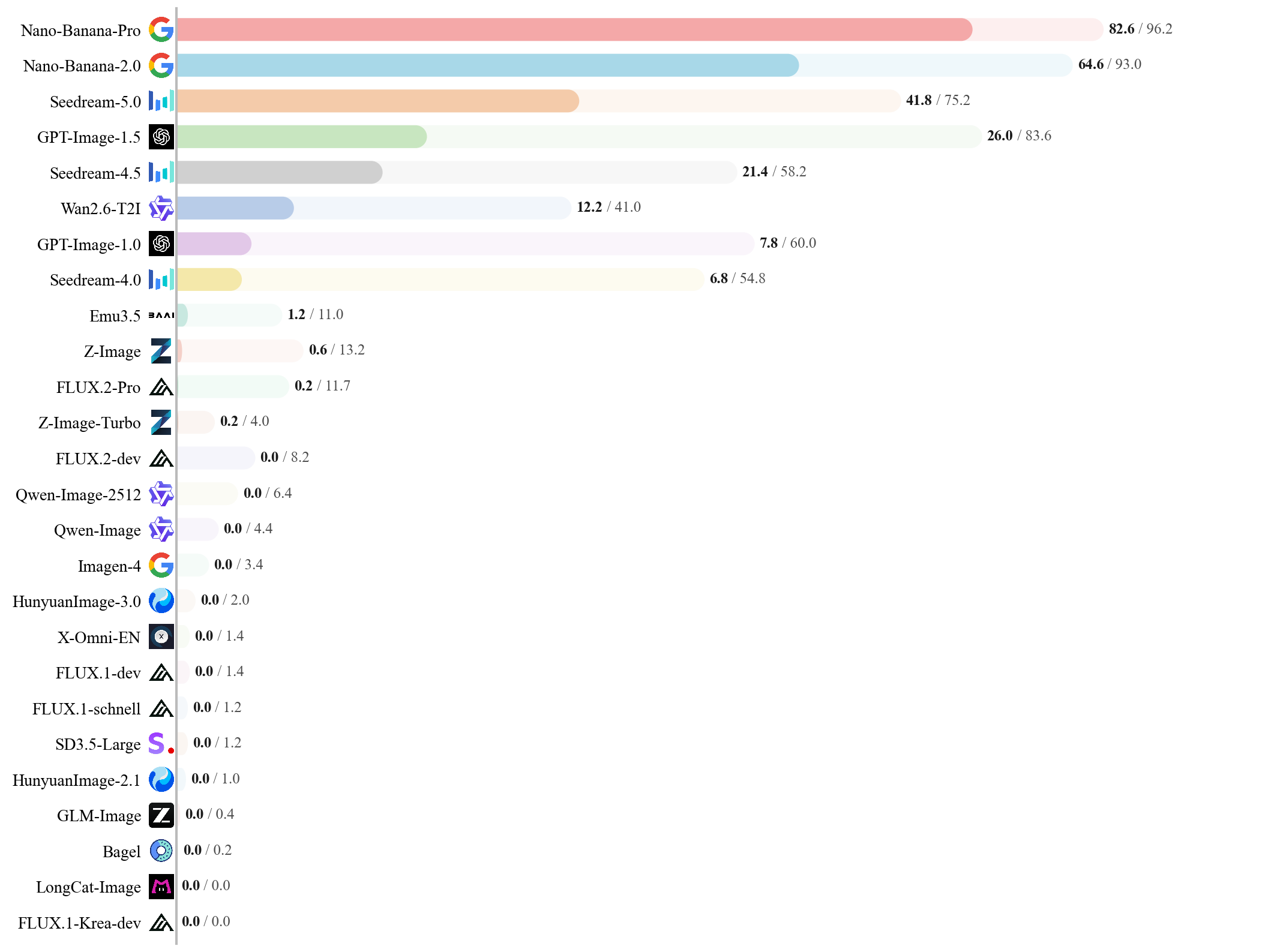}
    \caption{\textbf{Capability Ladder: Knowledge-based Reasoning.}}
    \label{fig:rank_know_cap}
\end{figure}

To understand the underlying technical strengths of the models, we present results across four fundamental dimensions, including Layout Control (Fig.~\ref{fig:rank_layout_cap}), Attribute Binding (Fig.~\ref{fig:rank_attr_cap}), Text Rendering (Fig.~\ref{fig:rank_text_cap}), and Knowledge-based Reasoning (Fig.~\ref{fig:rank_know_cap}).

\section{Evaluator Stability Analysis}
\label{sec:evaluator_reliability}

\subsection{Human-Alignment and Case Studies}
As shown in Table~\ref{tab:human_alignment}, the selected MLLM judge exhibits a high degree of correlation with expert human annotations. Our primary evaluator, Gemini-3-Flash, achieves the highest agreement and Cohen's Kappa, significantly outperforming other models in the checklist-based protocol.

\begin{figure}[ht]
    \centering
    \includegraphics[width=\linewidth]{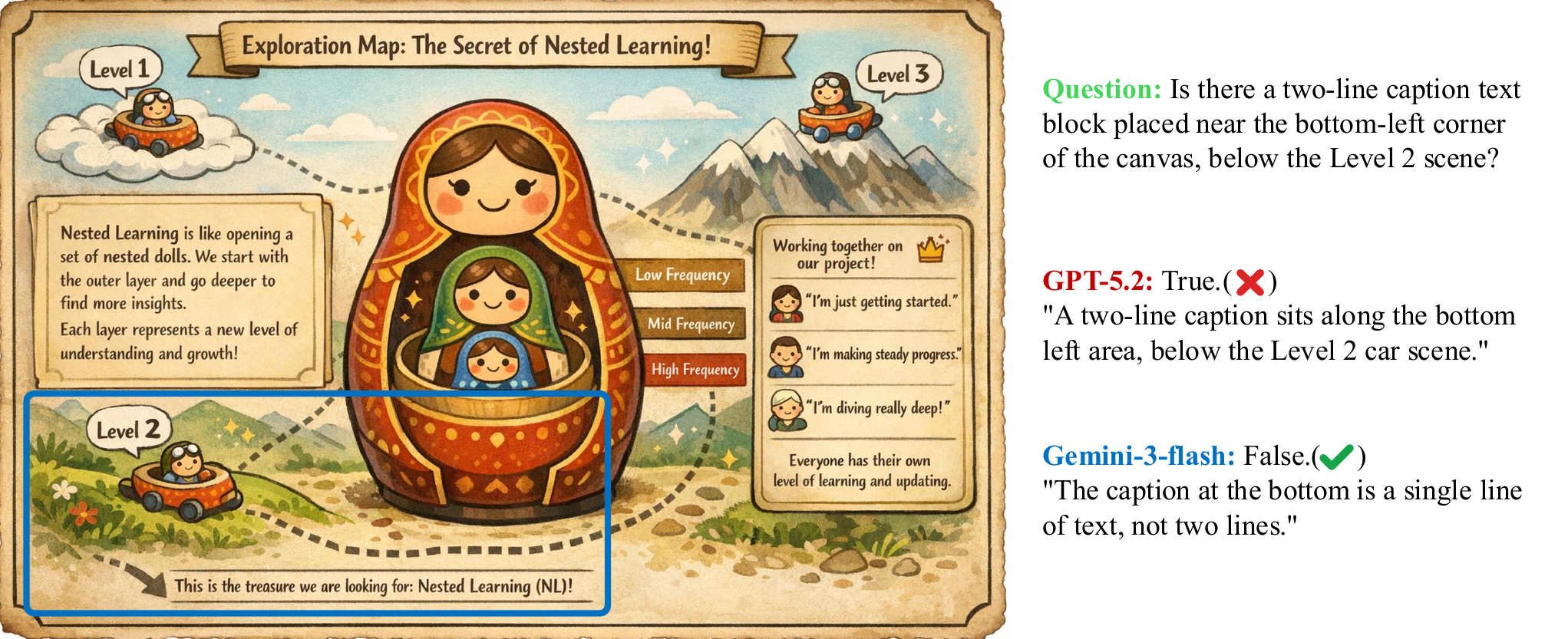}
    \caption{\textbf{Evaluator Case Study A.} Qualitative comparison of visual grounding performance between different MLLM judges.}
    \label{fig:judge_case1}
\end{figure}

\begin{figure}[ht]
    \centering
    \includegraphics[width=\linewidth]{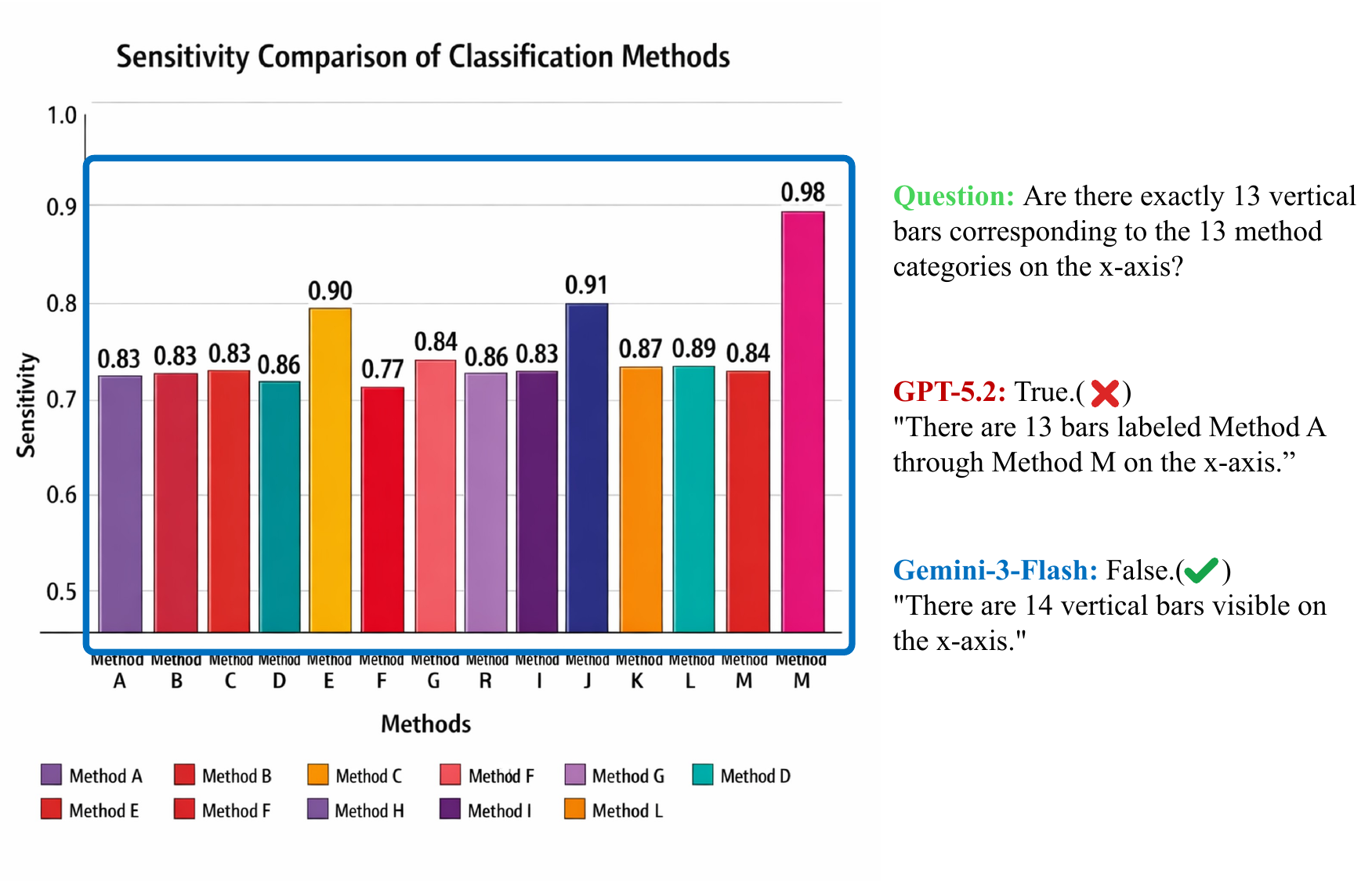}
    \caption{\textbf{Evaluator Case Study B.}}
    \label{fig:judge_case2}
\end{figure}

Figures~\ref{fig:judge_case1}-\ref{fig:judge_case2} further provide qualitative comparisons between different MLLM judges. These cases illustrate the superior visual grounding and reasoning depth of the selected evaluator. For instance, the judge accurately verifies complex spatial constraints, such as the placement of a two-line caption block below specific scene levels (Fig.~\ref{fig:judge_case1}), and performs precise element counting, such as verifying the exact presence of 13 vertical bars against 13 categories on a chart's x-axis (Fig.~\ref{fig:judge_case2}). Such detailed rationales demonstrate the robustness of using \textit{state-of-the-art} MLLMs to identify fine-grained errors and spatial inconsistencies that baseline judges often overlook.

\begin{table}[ht]
    \centering
    \caption{\textbf{Human-Judge Alignment Results.} Comparison of overall agreement ($p_o$) and Cohen's Kappa ($\kappa$) across different MLLM judges.}
    \label{tab:human_alignment}
    \begin{tabular}{c|cc}
        \toprule
       Evaluator Model  &  $p_o$ & $\kappa$ \\ \hline
        Gemini-3-Flash & 90.88\% & 0.7692 \\
        OpenAI-GPT-5.2 & 84.15\% & 0.6221 \\
        OpenAI-GPT-5.1 & 82.91\% & 0.5646 \\ \bottomrule
    \end{tabular}
\end{table}

\subsection{Stability Analysis}
We further verify the stability of our protocol in Table~\ref{tab:eval_variance} using Gemini-3-Flash as the judge. The minimal variance observed across three independent evaluation passes confirms that the checklist-based scoring system is stable.

\begin{table}[ht]
\centering
\caption{\textbf{Evaluator Stability Test.} Variance of results across three independent evaluation trials for selected models.}
\label{tab:eval_variance}
\begin{tabular}{lcccc}
\toprule
Model & Trial 1 & Trial 2 & Trial 3 & Std. Dev ($\sigma$) \\
\midrule
Nano-Banana-Pro & \scor{93.7}{76.7} & \scor{92.8}{76.1} & \scor{92.7}{76.1} & \scor{0.45}{0.28} \\
GPT-Image-1.5 & \scor{81.6}{35.9} & \scor{80.6}{35.9} & \scor{80.4}{35.8} & \scor{0.68}{0.05} \\
\bottomrule
\end{tabular}
\end{table}

\end{document}